\pdfoutput=1

\documentclass[11pt]{article}

\usepackage[preprint]{acl}

\usepackage{times}
\usepackage{latexsym}

\usepackage[T1]{fontenc}

\usepackage[utf8]{inputenc}

\usepackage{microtype}

\usepackage{inconsolata}

\usepackage{graphicx}

\usepackage[utf8]{inputenc} 
\usepackage[T1]{fontenc}    
\usepackage{hyperref}       
\usepackage{url}            
\usepackage{booktabs}       
\usepackage{amsfonts}       
\usepackage{nicefrac}       
\usepackage{microtype}      
\usepackage{xcolor}         
\usepackage{times}
\usepackage{latexsym}
\usepackage{hyperref}
\usepackage{hwemoji}
\usepackage{color, colortbl}
\definecolor{Gray}{gray}{0.9}
\usepackage{array}
\usepackage{booktabs}
\usepackage{multirow}
\usepackage{tcolorbox}
\usepackage{amsmath}
\usepackage{amssymb}
\usepackage{enumerate}
\usepackage{diagbox}
\usepackage{mathtools}
\usepackage{csquotes}
\usepackage{paralist}
\usepackage{tabularx}
\usepackage{lipsum}
\usepackage{ragged2e}
\usepackage{wrapfig}

\usepackage{inconsolata}

\newcommand{\name}[1]{\textsc{Mars}}

\newcommand{\emoji}{\raisebox{-4.5pt}{\includegraphics[width=1.2em]{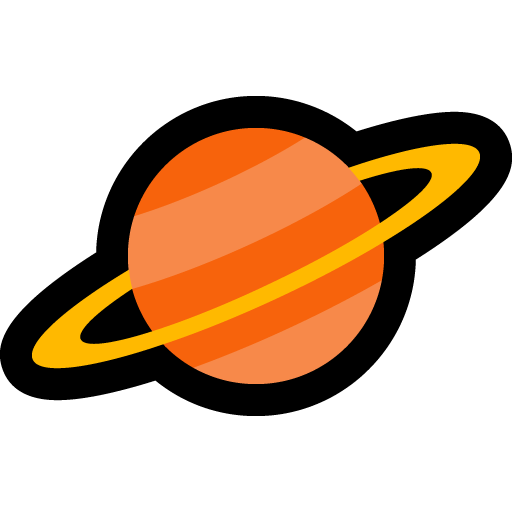}}}
\newcommand{\emojiname}{\raisebox{-2.5pt}{\includegraphics[width=1.1em]{Figures/mars.png}}\textsc{Mars}}
\newcommand{\original}[1]{\textit{\textcolor[HTML]{808080}{(#1)}}}

\definecolor{colorxmark}{RGB}{255, 87, 51}
\definecolor{colorcmark}{RGB}{66, 154, 137}
\definecolor{headcolor}{HTML}{018161}
\definecolor{relationcolor}{HTML}{d95f02}
\definecolor{tailcolor}{HTML}{6560a3}
\definecolor{concept_color}{HTML}{385723}
\definecolor{instance_color}{HTML}{1F4E79}
\definecolor{new_knowledge_color}{HTML}{7030A0}
\definecolor{original_tail_color}{HTML}{BF9000}

\title{\emoji\name{}: Benchmarking the Metaphysical Reasoning Abilities of Language Models with a Multi-task Evaluation Dataset}

\author{
Weiqi Wang, Yangqiu Song\\
Department of Computer Science and Engineering, HKUST, Hong Kong SAR, China\\
\texttt{\{wwangbw, yqsong\}@cse.ust.hk}\\ 
}

\begin{document}
\maketitle
\begin{abstract}
To enable Large Language Models (LLMs) to function as conscious agents with generalizable reasoning capabilities, it is crucial that they possess the ability to comprehend \textit{situational changes (transitions) in distribution} triggered by environmental factors or actions from other agents. 
Despite its fundamental significance, this ability remains underexplored due to the complexity of modeling infinite possible changes in an event and their associated distributions, coupled with the lack of benchmark data with situational transitions.
Addressing these gaps, we propose a novel formulation of \textit{reasoning with distributional changes} as a \textit{three-step discriminative process}, termed as \textit{\textbf{\underline{M}et\underline{A}physical \underline{R}ea\underline{S}oning}}.
We then introduce the first-ever benchmark,~\emojiname{}, comprising three tasks corresponding to each step.
These tasks systematically assess LLMs' capabilities in reasoning the plausibility of (i) changes in actions, (ii) states caused by changed actions, and (iii) situational transitions driven by changes in action.
Extensive evaluations with 20 (L)LMs of varying sizes and methods indicate that all three tasks in this process pose significant challenges, even after fine-tuning.
Further analyses reveal potential causes for the underperformance of LLMs and demonstrate that pre-training on large-scale conceptualization taxonomies can potentially enhance LMs' metaphysical reasoning capabilities.
Our data and models are publicly accessible at \href{https://github.com/HKUST-KnowComp/MARS}{https://github.com/HKUST-KnowComp/MARS}.
\end{abstract}

\section{Introduction}

\begin{figure}[t]
     \centering
     \includegraphics[width=1\linewidth]{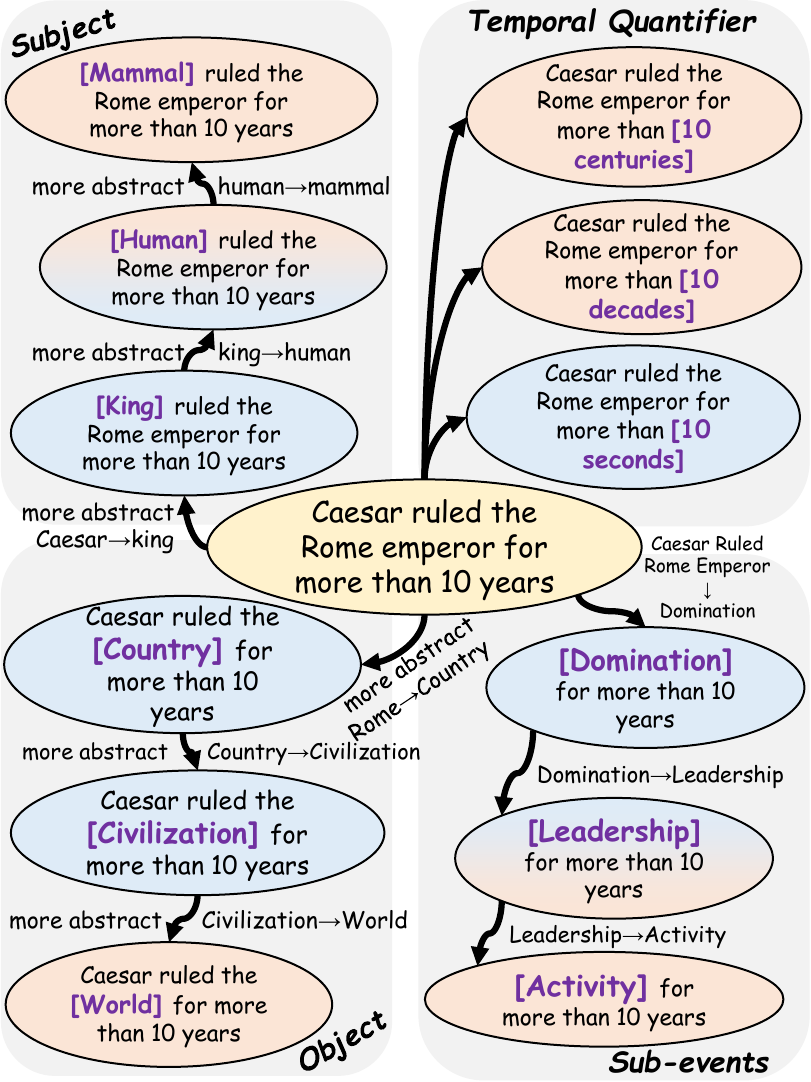}
     \vspace{-0.25in}
     \caption{Examples of changes in event in our formulation. After changes occur, events may become \textbf{\textcolor[HTML]{F4B183}{metaphysical}} as components are abstracted into high-level concepts, while some remain plausible in reality.}
    \label{fig:change_example}
    \vspace{-0.3in}
\end{figure}

Recent advances in LLMs have demonstrated superior performance in a variety of reasoning tasks~\cite{DBLP:conf/acl/LiuYZF023,DBLP:conf/eacl/ChanCWJFLS24,DBLP:conf/emnlp/KoLKRK23,DBLP:conf/emnlp/QinZ0CYY23,DBLP:conf/emnlp/JainSA0JD23}.
However, to truly achieve conscious processing~\cite{DBLP:conf/emnlp/Andreas22}, the integration of System II reasoning ability~\cite{sloman1996empirical,daniel2017thinking} is essential as it enables LLMs to perform out-of-distribution generalization when encountered with unfamiliar scenarios~\cite{DBLP:journals/cacm/BengioLH21}.
Among several components that make up System II reasoning, a critical element of it is the ability to \textit{reason with situational changes in distribution}, triggered by \textit{environmental factors} and \textit{actions by themselves or other agents}, when dealing with non-stationarities~\cite{DBLP:journals/corr/abs-1709-08568}.
It serves as the core ability in planning tasks~\cite{DBLP:journals/corr/abs-2402-02716}, which can be achieved by dynamically recombining existing concepts in the given environment or action and learning from the resultant situational changes~\cite{DBLP:conf/icml/LakeB18,DBLP:conf/iclr/BahdanauMNNVC19,DBLP:conf/nips/VriesBMCB19}.
For instance, in the event that ``PersonX is driving a car in a sunny day,'' a change in the weather from sunny to rainy could cause a different outcome, such as ``PersonX becomes more cautious and drives slower.''
This illustrates that a change in weather conditions can lead to a change in the driver's behavior, which represents an environmental change that triggers situational changes within the distribution of different weathers.

Though fundamental, the exploration of this ability has been limited due to several factors. 
First, the scope for change within an event is vast, with numerous components capable of altering in a wide variety of ways. 
This results in an overwhelmingly large number of potential changes that are impossible to fully cover with existing knowledge bases. 
Second, \textit{reasoning with changes in distribution} lacks a clear formulation due to its complexity. 
Unlike one-step inference reasoning tasks~\cite{DBLP:conf/aaai/SapBABLRRSC19}, changes in action may lead to implausible events that cannot occur in reality, thus terminating the reasoning process. 
Such type of changes require extra care when designing evaluation protocols.
Lastly, there is a lack of a reliable evaluation benchmark. 
Existing benchmarks~\cite{DBLP:conf/nips/ValmeekamMHSK23,DBLP:conf/acl/0001HX023} typically focus on a limited number of changes within a few scenarios, thus limiting the coverage of formed distributions. 
The changes in actions and states are also formulated under planning or logical tasks, which neglect transitions (consequences) caused by changes.

To address these gaps, we take a step forward by formally defining \textit{reasoning with changes in distribution} as a \textit{three-step discriminative process}. 
We start by defining seven categories of changes, each corresponding to different components within an event. 
To semantically cover more changes in a unified manner, we propose implementing changes by altering each component within the event using their abstractions or numerical variations. 
This approach creates a hierarchical distribution of various changes, with the abstracted ones offering a more generalized coverage.
Inspired by~\citet{DBLP:journals/cacm/BengioLH21}, we formulate \textit{reasoning with changes in distribution} as sequentially tasking the model to: (1) assess the plausibility of a potential change in a given event that describes an action, (2) evaluate the plausibility of an inferential state resulting from the modified action, and (3) determine the necessary change in an action to convert an implausible inferential state into a plausible one. 
We refer to this process as \textit{\textbf{metaphysical reasoning}}--a term we adopt to describe a mode of reasoning that deals with highly improbable or abstract scenarios distinct from its traditional philosophical meaning or counterfactual reasoning (see Appendix~\ref{appendix:metaphysical_versus_counterfactual})--as it also requires models to distinguish implausible actions, states, and transitions that exist only in this abstract ``metaphysical'' realm, indicating their rare occurrence in reality~\cite{heidegger2014introduction}.

We then construct the first evaluation benchmark,~\emojiname{}, featuring 355K annotated data across three tasks corresponding to each step.
It is constructed by sequentially instructing an LLM to extract events from Wikitext~\cite{DBLP:conf/iclr/MerityX0S17} and BookCorpus~\cite{DBLP:conf/iccv/ZhuKZSUTF15}, identify mutable components within each event, generate abstractions and numerical variations for those components, create a metaphysical inference state based on the changes, and generate the necessary modifications to make the metaphysical inference plausible in reality. 
Large-scale human annotations are then conducted to provide labels of evaluation data entries and verify the quality of our benchmark. 
Extensive experiments with over 20 (L)LMs demonstrate that all three tasks in this process present significant challenges, even for LMs after fine-tuning. 
Further analyses reveal potential reasons for such underperformance and identify possible solutions for enhancing the metaphysical reasoning abilities of language models.


\section{Backgrounds and Related Works}
\textbf{Reasoning about Changes in Distribution.}
Enabling LMs to understand distributional changes due to localized causal interventions, particularly in semantic spaces, has long been a crucial objective in the pursuit of conscious machine intelligence~\cite{bengio2019system,DBLP:journals/cacm/BengioLH21}.
Previous works have mainly explored this within the context of discriminating changes between actions and states with methods such as commonsense knowledge injection~\cite{DBLP:conf/emnlp/TandonDGYBC18}, event calculus~\cite{DBLP:books/sp/22/BasinaPP22}, and fuzzy reasoning~\cite{DBLP:journals/ijis/ZhangLS13}. 
Other studies aim to benchmark this reasoning process through logical reasoning tasks~\cite{DBLP:conf/acl/0001HX023} and planning tasks~\cite{DBLP:conf/nips/ValmeekamMHSK23,DBLP:conf/nips/WuYC0G21}. 
However, these studies only cover changes in limited formats and scenarios and also overlook the significance of representing changes as a distribution in relation to different variables in actions. 
Such loss restricts the out-of-distribution generalizability of the resulting LMs when facing unfamiliar scenarios. 
Moreover, previous evaluations do not cover transitions caused by changes, making subsequent evaluations around reasoning with changes incomplete.

\begin{figure*}[t]
     \centering
     \includegraphics[width=1\linewidth]{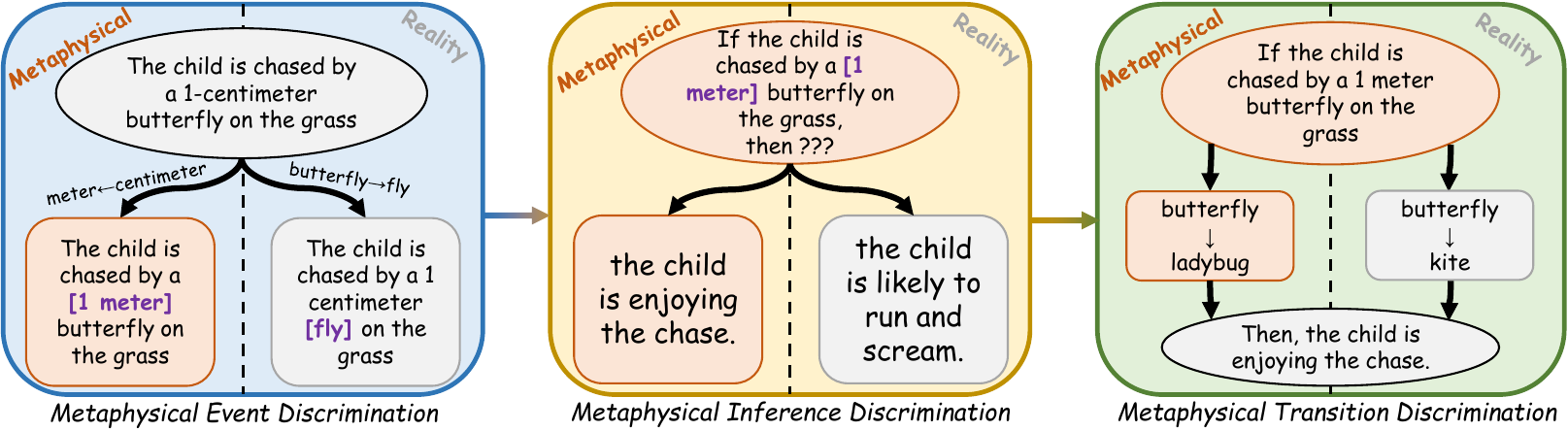}
     \caption{The three steps in metaphysical reasoning. 
     Our motivation behind this is that, by conquering all steps sequentially, a conscious agent could answer: (1) Will the change occur in reality? (2) What will the change cause? (3) What change can make a \textbf{\textcolor[HTML]{F4B183}{metaphysical}} (desired) inference plausible?     
     }
    \label{fig:tasks}
\end{figure*}

\noindent\textbf{Benchmarking LLMs.}
The advent of LLMs~\cite{openai2022chatgpt,GPT4,DBLP:journals/corr/abs-2307-09288,DBLP:journals/corr/abs-2302-13971,DBLP:journals/corr/abs-2403-05530} has sparked various studies in investigating LLM's potential in a variety of tasks~\cite{DBLP:conf/aaai/ChenXWLM24,DBLP:conf/aaai/0011LH024,DBLP:conf/www/YuanXHA24,DBLP:conf/eacl/ChanCWJFLS24,DBLP:conf/emnlp/JainSA0JD23,DBLP:conf/emnlp/QinZ0CYY23}.
These studies have significantly contributed to our understanding of LLMs by evaluating their performance across diverse tasks, using different scales of parameters and prompting methods~\cite{DBLP:conf/acl/QiaoO0CYDTHC23}. 
However, there is an absence of a comprehensive benchmark for assessing the ability of (L)LMs to \textit{reason with changes in distribution}. 
This inspires us to formally define it and introduce the first benchmark that evaluates such reasoning capabilities of (L)LMs.

\section{Definitions of Changes in Event and Metaphysical Reasoning}
\label{sec:task_definitions}
Modeling changes within an event is inherently complex due to the infinite number of changes that can occur. 
For simplicity, we only consider events that represent an action and study changes between their inferential states.
Given an event $e$, we first define seven types of changes that could transpire within $e$. 
These changes are represented as components of the event, including its subject $s$, verb $v$, object $o$, temporal quantifier $t$, spatial quantifier $l$, numerical properties $n$, and sub-events $se$.
The original event is denoted as a function of these seven components, $e=f(s,v,o,t,l,n,se)$. 
A change in the event can be represented by altering one of its components, for instance, $e'=f(s',v,o,t,l,n,se)$ if the change impacts the subject $s'$.

To effectively model the distribution of changes across different types of components, we leverage two types of hierarchical formulations.
Specifically, for $s, v, o, se$, we define changes in these components as conceptualizing their original instance into three concepts with progressively increased abstractedness~\cite{giunchiglia1992theory,tenenbaum2011grow}.
For $t,l,n$, we define their changes as modifications from their original values to three distinct numerical or spatial values with progressively increased units.
This brings a hierarchical structure to changes of a certain component, forming a distribution that gradually covers more possible changes. 
Abstracted components, as high-level concepts, can semantically represent a broader range of combinations for altering an event.
Some running examples of how changes impact an action are shown in Figure~\ref{fig:change_example}.
We then propose a \textit{three-step discriminative process}, which we term as \textbf{\textit{Metaphysical Reasoning}} (see Appendix~\ref{appendix:metaphysical_versus_counterfactual}), to formulate \textit{reason with changes in distribution}. 
The three steps, as shown in Figure~\ref{fig:tasks}, are:

\noindent\textbf{(1) Metaphysical Event Discrimination:}  
The first step answers the question, ``Will the change happen in reality?''
It aims to determine the plausibility of a change based on a given event, as alterations in components may lead to implausible events that defy reality. 
We refer to such an event, which rarely occurs in reality due to these changes, as a \textbf{\textit{metaphysical event}}. 
The goal of the first task is to discriminate whether the modified event $e'$, conditioned on the original event e with a single altered component $c\in(s,v,o,t,l,n,se)$, is metaphysical or not by making a binary prediction.

\noindent\textbf{(2) Metaphysical Inference Discrimination:} 
Considering that distributional changes occur in non-stationary environments, a conscious agent should be able to predict the potential outcomes of the modified event for future reasoning scenarios. 
Therefore, the second step aims to answer the question, ``What will the altered event result in?'' 
Similarly, we term the inferences of an event that rarely occurs in reality as \textbf{\textit{metaphysical inference}}. 
The objective of the second task is to determine whether an inferential state $i$, triggered by the altered event $e'$, is metaphysical or not by predicting a binary answer. 
Note that $e'$ could be either metaphysical or not, as inferences in both cases can be evaluated.

\noindent\textbf{(3) Metaphysical Transition Reasoning:} 
Finally, with some inferences remain metaphysical, a conscious agent should be able to plan what change is necessary to make such inference plausible in reality. 
This completes the reasoning chain by covering the feasibility, consequence, and motivation of distributional changes. 
Thus, the last task answers the question, ``What change is needed to make a metaphysical inference plausible?''
We refer to this as \textbf{\textit{metaphysical transition reasoning}} and set the objective as to determine whether another change, denoted as $c'$, can make a metaphysical inference $i$ plausible in relation to a changed event $e'$ by making a binary prediction regarding $c'$.

\section{\emojiname{} Benchmark Curation Pipeline}
We then introduce our sequential pipeline for curating the~\emojiname{} benchmark.
An overview of our curation pipeline is shown in Appendix Figure~\ref{fig:data_curation_pipeline}.
To guarantee a comprehensive coverage of events across various domains and topics, we source original text from two publicly available large corpora: Wikitext~\cite{DBLP:conf/iclr/MerityX0S17} and BookCorpus~\cite{DBLP:conf/iccv/ZhuKZSUTF15}. 
We filter out noisy text that includes hashtags and hyperlinks and segment long text into sentences with no more than 200 tokens to facilitate future processing.

\subsection{Text Decomposition and Extraction}
\label{sec:text_decomposition}
We first perform text decomposition~\cite{DBLP:conf/sigir/YeHYLHL23,DBLP:journals/corr/abs-2305-08677} to break down lengthy text into semantically complete short events, which are then used for fine-grained component extraction. 
To enable large-scale processing, we use ChatGPT~\cite{openai2022chatgpt}, a powerful LLM with strong text understanding abilities, as the core processor for all stages. 
For each stage, we guide it with a few-shot prompt~\cite{DBLP:conf/naacl/WestBHHJBLWC22,DBLP:conf/nips/BrownMRSKDNSSAA20} by creating task-specific explanations and exemplars (detailed prompts are in Appendix~\ref{appendix:prompts}):
\begin{center}
\resizebox{\linewidth}{!}{
\begin{tabular}{l}
\textbf{\texttt{<TASK-PROMPT>}}\\
\textbf{\texttt{\textcolor{headcolor}{<INPUT$_1$>}\textcolor{tailcolor}{<OUTPUT$_{(1,1)}$>}}}~~\ldots~~\textbf{\texttt{\textcolor{tailcolor}{<OUTPUT$_{(1,N_1)}$>}}}\\
\textbf{\texttt{\textcolor{headcolor}{<INPUT$_2$>}\textcolor{tailcolor}{<OUTPUT$_{(2,1)}$>}}}~~\ldots~~\textbf{\texttt{\textcolor{tailcolor}{<OUTPUT$_{(2,N_2)}$>}}}\\
\ldots \\
\textbf{\texttt{\textcolor{headcolor}{<INPUT$_{10}$>}\textcolor{tailcolor}{<OUTPUT$_{(10,1)}$>}}}~~\ldots~~\textbf{\texttt{\textcolor{tailcolor}{<OUTPUT$_{(10,N_{10})}$>}}}\\
\textbf{\texttt{\textcolor{headcolor}{<INPUT$_{11}$>}}}
\end{tabular}
}
\end{center}
To perform text decomposition, \textbf{\texttt{<TASK-PROMPT>}} clarifies the goal to ChatGPT, which involves extracting semantically complete actions from the given text. 
\textbf{\texttt{\textcolor{headcolor}{<INPUT$_{1-10}$>}}} and \textbf{\texttt{\textcolor{tailcolor}{<OUTPUT$_{1-10}$>}}} are filled with 10 pairs of human-crafted examples, each containing several action events extracted from text sampled from Wikitext and BookCorpus. 
ChatGPT is expected to learn from these examples and use them as a guide to extract action events (\textbf{\texttt{\textcolor{tailcolor}{<OUTPUT$_{(11,1-N)}$>}}}) from the final input text (\textbf{\texttt{\textcolor{headcolor}{<INPUT$_{11}$>}}}).
For component extraction, we adjust \textbf{\texttt{<TASK-PROMPT>}} to define the task of extracting the seven components from a given event. 
We populate \textbf{\texttt{\textcolor{headcolor}{<INPUT$_{1-10}$>}}} and \textbf{\texttt{\textcolor{tailcolor}{<OUTPUT$_{1-10}$>}}} with 10 pairs of events and seven comma-separated lists of components extracted from the event, each corresponding to one type of components defined in \S\ref{sec:task_definitions}. 
ChatGPT then extracts seven lists of components for the final given event (\textbf{\texttt{\textcolor{headcolor}{<INPUT$_{11}$>}}}).
If any type of component is absent, ``None'' will be generated instead.

\subsection{Component Abstraction and Variation}
\label{sec:component_abstraction_variation}
The next step is designed to implement changes within the event by altering its components, extracted from the previous step, by generating their abstractions or numerical variations. 
Following~\citet{DBLP:journals/corr/abs-2401-07286}, we guide ChatGPT by modifying~\textbf{\texttt{<TASK-PROMPT>}} with the objective of generating abstract concepts for $s,v,o,se$ and numerical variations for $t,l,n$ within a specified event. 
For each \textbf{\texttt{\textcolor{headcolor}{<INPUT$_{1-10}$>}}} and \textbf{\texttt{\textcolor{tailcolor}{<OUTPUT$_{1-10}$>}}} pair, we populate the input with a specific event and one of its components. 
The output consists of three human-authored component abstractions or numerical variations that align with the event's context.
Subsequently, ChatGPT is tasked with generating three abstractions or numerical variations for the final pair of the given event and a component within the event (\textbf{\texttt{\textcolor{headcolor}{<INPUT$_{11}$>}}}).
Replacing the original components in the event with their generated changes forms changed event candidates for the metaphysical event discrimination task.

\begin{table*}[t]
\small
\centering
\resizebox{1\linewidth}{!}{
\begin{tabular}{@{}l|ccc|ccccc|c@{}}
\toprule
Dataset / Task & \#Text & \#Event & \#Avg.Token & \#Train & \#Dev & \#Test & \#Total. & \#Unlabel. & Expert. \\ 
\midrule
AbsATM~\cite{HE2024104149} & N/A & 7,196 & 1.060 & 107,384 & 12,117 & 11,503 & 131,004 & 372,584 & N/A \\
AbsPyramid~\cite{DBLP:journals/corr/abs-2311-09174} & N/A & 16,944 & 1.690 & 176,691 & 22,050 & 22,056 & 220,797 & 0 & N/A \\
\textbf{Meta. Event.} & 9,998 & 55,190 & 1.040 & 96,004 & 12,013 & 11,982 & 119,999 & 329,540 & 94.0\% \\
\midrule
AbsATM~\cite{HE2024104149} & N/A & 7,196 & 6.413 & 65,386 & 8,403 & 7,408 & 81,197 & 5,921,195 & N/A \\
\textbf{Meta. Inference.} & 9,837 & 35,528 & 10.40 & 96,009 & 12,010 & 11,981 & 120,000 & 497,590 & 96.5\% \\
\midrule
Propara~\cite{DBLP:conf/naacl/DalviHTYC18} & 9,051 & 9,051 & N/A & 7,043 & 913 & 1,095 & 9,051 & 0 & N/A \\
TRAC~\cite{DBLP:conf/acl/0001HX023} & 15,000 & 15,000 & N/A & 10,000 & 2,000 & 3,000 & 15,000 & 0 & N/A \\
PlanBench~\cite{DBLP:conf/nips/ValmeekamMHSK23} & 26,250 & 26,250 & N/A & 0 & 0 & 26,250 & 26,250 & 0 & N/A \\
\textbf{Meta. Transition.} & 9,677 & 31,447 & 1.810 & 92,495 & 11,563 & 11,560 & 115,618 & 273,474 & 93.5\% \\
\bottomrule
\end{tabular}
}
\caption{Statistics of the~\name{} benchmark in comparison against other benchmarks.
Meta. refers to three tasks in~\name{}.
Expert. refers to expert verification results.}
\label{tab:mars_statistics}
\end{table*}

\subsection{Inference Generation}
\label{sec:inference_generation}
We then collect inferential states of the modified events by similarly instructing ChatGPT to autonomously generate them. 
For each altered event, we prompt ChatGPT to separately generate one plausible inference and one metaphysical inference. 
We first modify \textbf{\texttt{<TASK-PROMPT>}} to generate a state that could potentially be caused by the altered event, and populate \textbf{\texttt{\textcolor{headcolor}{<INPUT$_{1-10}$>}}} with 10 modified events and \textbf{\texttt{\textcolor{tailcolor}{<OUTPUT$_{1-10}$>}}} with 10 corresponding plausible inferences authored by human experts. 
ChatGPT is then requested to generate an additional plausible state inference for the given changed event (\textbf{\texttt{\textcolor{headcolor}{<INPUT$_{11}$>}}}).
Next, we adjust \textbf{\texttt{<TASK-PROMPT>}} to generate a metaphysical state that is infrequently caused by the changed event in reality, yet remains contextually relevant. 
We replace \textbf{\texttt{\textcolor{tailcolor}{<OUTPUT$_{1-10}$>}}} with 10 metaphysical inferences and then collect a metaphysical inference from ChatGPT. 
This, along with the generated plausible inference, forms two candidate data entries for each changed event in the metaphysical inference discrimination task.

\subsection{Metaphysical Transition Generation}
\label{sec:transition_reasoning}
Given that half of the inferential states generated in the previous step remain metaphysical, we then collect the additional changes necessary to transform these states into plausible real-world inferences. 
We adjust the \textbf{\texttt{<TASK-PROMPT>}} to describe such required changes and populate \textbf{\texttt{\textcolor{headcolor}{<INPUT$_{1-10}$>}}} with 10 pairs of modified events and their corresponding metaphysical inferences. 
\textbf{\texttt{\textcolor{tailcolor}{<OUTPUT$_{1-10}$>}}} are filled with 10 corresponding human-authored changes in events that can render the inferences plausible. 
Subsequently, ChatGPT generates the required change for the final pair of the modified event and its metaphysical inference (\textbf{\texttt{\textcolor{headcolor}{<INPUT$_{11}$>}}}). 
Note that the generated change still needs to be one of the seven types we defined in \S\ref{sec:task_definitions}. 
We collect one additional change for each metaphysical inference and use it as a candidate data entry for the last task.
However, we discard event and inference pairs that ChatGPT deems impossible to render plausible, even with an additional change.

\begin{figure}[t]
     \centering
     \includegraphics[width=0.8\linewidth]{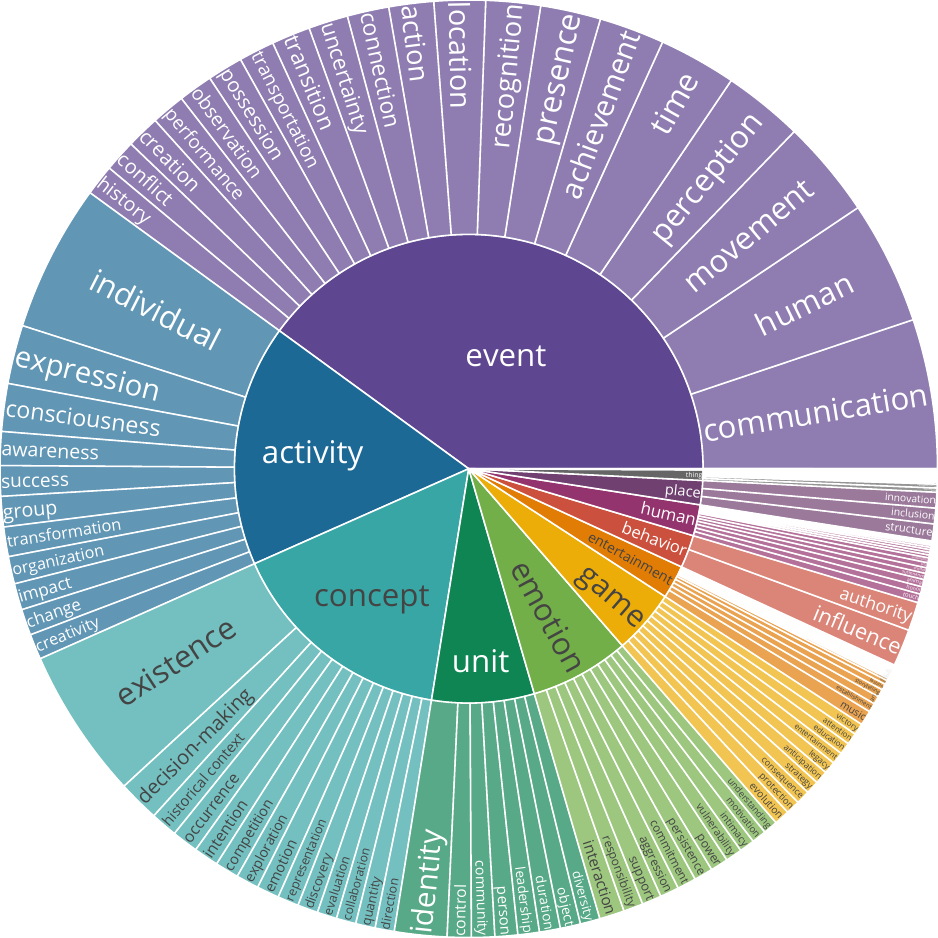}
     \caption{Hypernym distribution of the top 5,000 popular component variations.}
    \label{fig:hypernym}
\end{figure}

\subsection{Human Annotations}
\label{sec:human_annotation}
\textbf{Annotation:}
Finally, we carry out large-scale human annotations to label candidate data for each task via Amazon Mechanical Turk (AMT).
We provide detailed instructions with examples to qualified workers and task them with annotating (1) the plausibility of the changed events generated in~\S\ref{sec:component_abstraction_variation}, (2) the plausibility of the plausible/metaphysical inferences produced in~\S\ref{sec:inference_generation}, and (3) the plausibility of the transitions generated in~\S\ref{sec:transition_reasoning}.
We collect five votes for each entry and the majority vote is used as the final label.
The overall inter-annotator agreement (IAA) is 81\% in terms of pairwise agreement, and the Fleiss Kappa~\cite{fleiss1971measuring} is 0.56, indicating sufficient agreement (see Appendix~\ref{appendix:annotation_details}).

\noindent\textbf{Expert Verification:} 
To verify the quality of our collected labels, we recruit three postgraduate students with rich experience in NLP to perform a second round annotation. 
Each of them is asked to annotate a sample of 100 data entries for each task, following the same instructions provided to the AMT annotators. 
Results in Table~\ref{tab:mars_statistics} show that, on average, 93.67\% labels collected from human annotations align with the expert's vote, demonstrating the reliability of our collected labels.

\begin{table*}[t]
    \small
    \centering
    \resizebox{1\linewidth}{!}{
	\begin{tabular}{@{}llccccccccc@{}}
	\toprule
    \multirow{2}{*}{\textbf{Methods}}&\multirow{2}{*}{\textbf{Backbone}}&\multicolumn{3}{c}{\textbf{Event}} &\multicolumn{3}{c}{\textbf{Inference}}&\multicolumn{3}{c}{\textbf{Transition}}\\
    \cmidrule(lr){3-5}\cmidrule(lr){6-8}\cmidrule(lr){9-11}
	&&\textbf{Acc}&\textbf{AUC}&\textbf{Ma-F1}&\textbf{Acc}&\textbf{AUC}&\textbf{Ma-F1}&\textbf{Acc}&\textbf{AUC}&\textbf{Ma-F1}\\
            \midrule
            \textbf{Random} & \multicolumn{1}{c}{-} & 50.00 & - & 49.56 & 50.00 & - & 49.56 & 50.00 & - & 49.56 \\
            \textbf{Majority} & \multicolumn{1}{c}{-} & 60.98 & - & 37.99 & 58.56 & - & 36.93 & 50.25 & - & 33.37 \\
		  \midrule
            \multirow{6}{*}{\textbf{\begin{tabular}[c]{@{}l@{}}PTLM\\ \textit{(Zero-shot)}\end{tabular}}}
            &DeBERTa-Base \scriptsize{\textit{214M}} & \underline{60.55} & 49.41 & 42.89 & 50.10 & 47.57 & 48.96 & 49.05 & 41.32 & 33.19 \\
            &DeBERTa-Large \scriptsize{\textit{435M}} & 48.27 & 49.88 & 45.87 & 47.73 & 49.94 & 44.44 & 50.73 & 46.96 & 46.15 \\
            &GPT2-XL \scriptsize{\textit{1.5B}} & 38.62 & \underline{51.12} & 27.93 & 44.40 & 51.88 & 31.45 & 49.92 & 48.35 & 48.09 \\
            &CAR \scriptsize{\textit{435M}} & 54.63 & 49.34 & 49.96 & 48.33 & 42.85 & 41.93 & 52.97 & 35.05 & 46.94 \\
            &CANDLE \scriptsize{\textit{435M}} & 51.90 & 49.12 & \underline{50.30} & 46.77 & 44.03 & 38.48 & 53.49 & 34.95 & 47.95 \\
            &VERA \scriptsize{\textit{11B}} & 51.82 & 50.48 & 48.52 & \underline{60.97} & \underline{62.54} & \underline{59.09} & \underline{61.31} & \underline{66.32} & \underline{61.17} \\
            \midrule
            \multirow{4}{*}{\textbf{\begin{tabular}[c]{@{}l@{}}PTLM\\ \textit{(Fine-tuned)}\end{tabular}}}
            &DeBERTa-Base \scriptsize{\textit{214M}} & 63.82 & 63.98 & \textbf{\underline{63.39}} & 69.50 & 70.59 & 69.31 & 71.96 & 73.85 & 71.17 \\ 
            &DeBERTa-Large \scriptsize{\textit{435M}} & \textbf{\underline{64.45}} & \textbf{\underline{64.16}} & 63.27 & \textbf{\underline{69.57}} & \textbf{\underline{71.15}} & 69.33 & \textbf{\underline{72.93}} & 74.00 & 72.01 \\
            &GPT2-XL \scriptsize{\textit{1.5B}} & 46.68 & 47.63 & 46.96 & 43.70 & 44.22 & 30.41 & 44.57 & 45.03 & 45.89 \\ 
            &VERA \scriptsize{\textit{11B}} & 61.95 & 61.43 & 60.81 & 63.90 & 66.93 & \textbf{\underline{70.84}} & 71.75 & \textbf{\underline{74.57}} & \textbf{\underline{73.27}} \\
            \midrule
            \multirow{15}{*}{\textbf{\begin{tabular}[c]{@{}l@{}}LLM\\ \textit{(Zero-shot)}\end{tabular}}}
            & Meta-LLaMa-2-7B & 50.64 & - & 41.41 & 49.87 & - & 49.23 & 50.94 & - & 50.64 \\
            & Meta-LLaMa-2-13B & 51.50 & - & 49.48 & 50.81 & - & 50.57 & 50.81 & - & 50.80 \\
            & Meta-LLaMa-2-70B & 52.40 & - & 49.03 & 56.13 & - & 46.81 & 48.45 & - & 48.34 \\
            & Meta-LLaMa-3-8B & 50.62 & - & 49.12 & 51.33 & - & 50.98 & 51.95 & - & 51.07 \\
            & Meta-LLaMa-3-70B & 57.41 & - & 50.59 & 63.40 & - & 61.82 & 60.15 & - & 60.01 \\
            & Meta-LLaMa-3.1-8B & 51.01 & - & 50.27 & 52.13 & - & 51.29 & 52.35 & - & 52.09 \\
            & Meta-LLaMa-3.1-70B & 59.22 & - & 52.08 & 63.61 & - & 61.90 & 61.28 & - & 61.03 \\
            &\quad+RAG & \underline{61.21} & - & \underline{54.51} & \underline{66.38} & - & \underline{65.90} & 61.53 & - & 61.22 \\
            &\quad+Multi-Agent & 56.12 & - & 51.08 & 65.06 & - & 65.01 & \underline{62.54} & - & \underline{62.19} \\
            &\quad+Self-reflection & 57.94 & - & 53.17 & 63.91 & - & 63.51 & 60.92 & - & 60.77 \\
            & Meta-LLaMa-3.1-405B & 60.01 & - & 52.99 & 64.52 & - & 63.23 & 61.74 & - & 61.76 \\
            & Gemma-2-9B & 56.88 & - & 48.53 & 51.83 & - & 51.76 & 49.41 & - & 45.01 \\
            & Falcon-7B & 54.32 & - & 49.51 & 51.77 & - & 50.30 & 50.42 & - & 49.02 \\
            & Falcon-40B & 52.35 &  - & 50.36 & 49.67 & - & 49.38 & 50.27 & - & 50.22 \\
            & Mistral-7B & 49.90 & - & 48.94 & 50.23 & - & 50.06 & 51.75 & - & 51.75 \\
            \midrule
            \multirow{5}{*}{\textbf{\begin{tabular}[c]{@{}l@{}}LLM\\ \textit{(Fine-tuned)}\end{tabular}}}
            & Meta-LLaMa-2-7B & 60.10 & 59.90 & 59.00 & 63.51 & 66.44 & 62.55 & 66.06 & 70.38 & 65.12 \\
            & Meta-LLaMa-2-13B & 60.67 & 60.64 & 60.00 & 64.61 & 67.67 & 63.59 & 68.22 & 72.19 & 66.37 \\
            & Meta-LLaMa-3-8B & 60.06 & 60.54 & 59.58 & 65.76 & 67.88 & 65.72 & 69.83 & 74.59 & 68.74 \\
            & Gemma-2-9B & \underline{61.23} & \underline{61.25} & \underline{60.28} & \underline{69.24} & \underline{70.76} & \underline{69.00} & \underline{73.30} & \underline{76.91} & \underline{69.18} \\
            & Mistral-7B & 60.35 & 60.77 & 60.07 & 66.91 & 70.06 & 65.95 & 71.87 & 75.47 & 68.53 \\
            \midrule
            \multirow{8}{*}{\textbf{\begin{tabular}[c]{@{}l@{}}LLM\\ \textit{(API)}\end{tabular}}}
            &GPT4 & 53.90 & - & 53.45 & 51.20 & - & 50.95 & \underline{49.41} & - & \underline{49.33} \\
            &GPT4 (5-shots) & 49.85 & - & 49.58 & 51.47 & - & 51.30 & 48.88 & - & 48.73 \\
            &GPT4 (COT) & 51.28 & - & 50.73 & 51.49 & - & 51.35 & 47.62 & - & 47.58 \\
            &GPT4 (SC-COT) & 51.97 & - & 51.26 & 52.05 & - & 52.27 & 48.24 & - & 48.11 \\
            &GPT-4o-mini & 57.94 & - & 57.91 & 53.84 & - & 53.53 & 48.06 & - & 48.06 \\
            &\quad+RAG & \underline{59.99} & - & \underline{59.97} & \underline{54.54} & - & \underline{54.21} & 49.39 & - & 49.19 \\
            &\quad+Multi-Agent & 54.21 & - & 53.17 & 52.76 & - & 52.26 & 46.94 & - & 46.70 \\
            &\quad+Self-reflection & 56.89 & - & 55.21 & 53.22 & - & 53.20 & 48.51 & - & 48.45 \\
		\bottomrule
	\end{tabular}
    }
\vspace{-0.1in}
\caption{Evaluation results (\%) of various language models on the testing sets of~\name{}. The best performances within each method are \underline{underlined} and the best among all methods are \textbf{bold-faced}.} 
\label{tab:main_eval_test}
\vspace{-0.1in}
\end{table*}

\section{Evaluations and Analysis}
\subsection{\emojiname{} Statistics}
Table~\ref{tab:mars_statistics} presents statistics of the~\name{} benchmark, which comprises a total of 355,617 annotated data distributed across three tasks.
We partition the annotated data into training, development, and testing splits following an 8:1:1 ratio, ensuring there is no overlap of text and events between the different splits to preserve the evaluation's generalizability. 
On average, 1.04 tokens are generated to describe changes in action for the metaphysical event and transition discrimination tasks, while 10.4 tokens are used for inferences in the metaphysical inference discrimination task. 
To the best of our knowledge, we are the first in proposing such a triad of tasks concurrently within a single benchmark. 
To compare~\name{} with other datasets, we select those with analogous task objectives for each task and compare them individually. 
We find~\name{} tends to be significantly larger than other benchmarks, covering a broader range of events and providing training sets for evaluating the performance of fine-tuned models.
To further illustrate the diverse coverage of events and changes in~\name{}, we match each component variation against hypernyms in Probase~\cite{DBLP:conf/sigmod/WuLWZ12} and plot their distribution according to their number of occurrences in Figure~\ref{fig:hypernym}. 
Our results indicate that~\name{} covers over 170,000 hypernyms in Probase, spanning broad categories such as event, activity, concept, unit, etc.

\begin{table*}[t]
\small
\centering
\resizebox{1\linewidth}{!}{
\begin{tabular}{@{}ll|ccccccccc@{}}
\toprule
    \multirow{2}{*}{\textbf{Backbone}}&\multirow{2}{*}{\textbf{Training Data}}&\multicolumn{3}{c}{\textbf{Event}} &\multicolumn{3}{c}{\textbf{Inference}}&\multicolumn{3}{c}{\textbf{Transition}}\\
    \cmidrule(lr){3-5}\cmidrule(lr){6-8}\cmidrule(lr){9-11}
    &&\textbf{Acc}&\textbf{AUC}&\textbf{Ma-F1}&\textbf{Acc}&\textbf{AUC}&\textbf{Ma-F1}&\textbf{Acc}&\textbf{AUC}&\textbf{Ma-F1}\\
\midrule
\multirow{4}{*}{\textbf{\begin{tabular}[c]{@{}l@{}}DeBERTa\\ \scriptsize{\textit{435M}}\end{tabular}}} 
& Zero-shot & 58.27 & 49.88 & 45.87 & 47.73 & 49.94 & 44.44 & 50.73 & 46.96 & 46.15 \\
& CANDLE & 57.94 & 58.22 & 57.31 & 59.43 & 59.03 & 58.18 & 62.00 & 62.19 & 61.50 \\
& \name{} & 64.45 & 64.16 & 63.27 & 69.57 & 71.15 & 69.33 & 72.93 & 74.00 & 72.01 \\
& CANDLE + \name{} & \textbf{\underline{64.95}} & \textbf{\underline{64.27}} & \textbf{\underline{63.74}} & \textbf{\underline{71.85}} & \underline{73.32} & \underline{71.64} & \textbf{\underline{74.39}} & \underline{77.97} & \underline{73.30} \\
\midrule
\multirow{4}{*}{\textbf{\begin{tabular}[c]{@{}l@{}}VERA\\ \scriptsize{\textit{11B}}\end{tabular}}}& Zero-shot & 41.82 & 50.48 & 38.52 & 60.97 & 62.54 & 59.09 & 61.31 & 66.32 & 61.17 \\
& CANDLE & 57.81 & 57.24 & 56.77 & 56.59 & 56.08 & 55.25 & 59.79 & 59.88 & 59.19 \\
& \name{} & 61.95 & 61.43 & 60.81 & 63.90 & 66.93 & \underline{70.84} & 71.75 & 74.57 & 73.27 \\
& CANDLE + \name{} & \underline{62.21} & \underline{61.77} & \underline{61.17} & \underline{71.45} & \textbf{\underline{74.46}} & 67.61 & \underline{73.95} & \underline{77.35} & \textbf{\underline{78.26}} \\
\midrule
\multirow{4}{*}{\textbf{\begin{tabular}[c]{@{}l@{}}LLaMa-3\\ \scriptsize{\textit{8B}}\end{tabular}}}& Zero-shot & 50.62 & - & 49.12 & 51.33 & - & 50.98 & 51.95 & - & 51.07 \\
& CANDLE & 56.47 & 56.75 & 56.07 & 58.29 & 57.81 & 57.00 & 58.74 & 58.81 & 58.19 \\
& \name{} & 60.06 & 60.54 & 59.58 & 65.76 & 67.88 & 65.72 & 69.83 & 74.59 & 68.74 \\
& CANDLE + \name{} & \underline{60.93} & \underline{60.80} & \underline{60.12} & \underline{69.13} & \underline{70.84} & \textbf{\underline{72.12}} & \underline{74.09} & \textbf{\underline{79.38}} & \underline{71.42} \\
\bottomrule
\end{tabular}
}
\caption{Evaluation results (\%) of transfering knowledge from CANDLE to aid~\name{}.
The best performances among each method is \underline{underlined} and best ones among all methods are \textbf{bold-faced}.
}
\label{tab:transfer_Concept_to_MARs}
\end{table*}

\subsection{Main Evaluations on \emojiname{}}
\subsubsection{Task Setup and Model Selections}
We then experiment with a selection of (L)LMs to investigate their performances on our curated~\name{} benchmark. 
Accuracy, AUC, and Macro-F1 scores are used as evaluation metrics.

The evaluation of different models are categorized into three types:
\textbf{(1) \textsc{Zero-shot}:} 
We first evaluate several (L)LMs in a zero-shot manner. 
For small-sized Pre-Trained Language Models (PTLMs), we evaluate DeBERTa-v3~\cite{DBLP:conf/iclr/HeGC23}, GPT2~\cite{radford2019language}, CAR~\cite{DBLP:conf/emnlp/WangF0XLSB23}, CANDLE~\cite{DBLP:journals/corr/abs-2401-07286}, and VERA~\cite{DBLP:conf/emnlp/0010WWS0H23}, following the design of zero-shot question answering~\cite{DBLP:conf/aaai/MaIFBNO21}. 
For LLMs, we evaluate LLaMa2, LLaMa3, LLaMa3.1~\cite{DBLP:journals/corr/abs-2302-13971,DBLP:journals/corr/abs-2307-09288, LLAMA3}, Gemma~\cite{Gemma}, Falcon~\cite{Falcon}, and Mistral~\cite{DBLP:journals/corr/abs-2310-06825} using direct zero-shot prompting~\cite{DBLP:conf/emnlp/QinZ0CYY23}. 
\textbf{(2) \textsc{Finetuning}:}
We then assess the performance of (L)LMs when fine-tuned on the training set of~\name{}. 
For PTLMs, we fine-tune DeBERTa, GPT2-xl, and VERA. 
For LLMs, we fine-tune LLaMa2, LLaMa3, Gemma, and Mistral using LoRA~\cite{DBLP:conf/iclr/HuSWALWWC22}. 
\textbf{(3) \textsc{LLM API}:} 
Finally, we evaluate the performance of GPT-4~\cite{GPT4} and GPT-4o-mini~\cite{GPT4omini}, which represent proprietary LLMs, under zero-shot, five-shots, Chain-of-Thought prompting (COT;~\citealp{wei2022chain}), and Self-Consistent COT (SC-COT;~\citealp{wang2022self}) settings.
For LLaMa3.1-70B and GPT-4o-mini, we also test their performances with RAG~\cite{DBLP:journals/corr/abs-2312-10997}, Multi-agent Calibration~\cite{DBLP:journals/corr/abs-2404-09127}, and Self Reflection~\cite{DBLP:journals/tacl/PanSXNWW24}.
Please find implementation details in Appendix~\ref{appendix:implementation_details}, multi-task fine-tuning experiments in Appendix~\ref{appendix:multitask-finetuning}, and few-shot fine-tuning experiments in Appendix~\ref{appendix:few-shot}.

\vspace{-0.05in}
\subsubsection{Results and Analysis}
\vspace{-0.05in}
Evaluation results are reported in Table~\ref{tab:main_eval_test}. 
From the results, we observe that:
\textbf{(1) Most models exhibit subpar performance under the zero-shot setting.}
Among PTLMs, only VERA delivers acceptable results across all three tasks, while the rest significantly underperform.
Though models fine-tuned on commonsense knowledge and conceptualizations, such as CAR and CANDLE, show some improvement compared to their DeBERTa-v3-Large backbone, these performances are still unsatisfactory, even falling below the level of majority voting. 
For LLMs, improving training paradigms and increasing the number of parameters can indeed help achieve better performance.
Nevertheless, all models perform poorly across all tasks in~\name{}, emphasizing the difficulty of our tasks.
\textbf{(2) Fine-tuning only offers limited benefits.} With fine-tuning, all models improve significantly.
For example, DeBERTa-Large's accuracy increases by 16.18\%, 21.84\%, and 22.2\% on three tasks, respectively.
However, the best results for all tasks are still capped at around 74\%, indicating a shared difficulty and significant room for future enhancements.
One potential reason for this is that, since we split the data according to the source of text in Wikitext and BookCorpus, the distribution between different splits may differ significantly, as the domain and topics could be diverse from each other.
We also discuss the reasons for PTLMs' strong performance compared to LLMs after fine-tuning in Appendix~\ref{appendix:ptlms_vs_llms}.
\textbf{(3) The GPT series models underperform compared to other LLMs, and COT does not consistently aid performance.}
Surprisingly, GPT series models fall short when compared to open LLMs, such as LLaMa-3-70B. 
One possible explanation is that negative examples in~\name{} are sourced from ChatGPT's generation and are obtained via post-human annotation. 
This makes it challenging to discriminate as these negative examples contradict GPT's internal knowledge.
Advanced prompting methods only offer limited improvement in performances.

\subsection{Analysis}
\subsubsection{Transferring from Conceptualization}
Improving the performance of LLMs on~\name{} requires extensive fine-tuning on large-scale human-annotated data, making it non-trivial.
Since we observe that approximately 80\% of action changes are executed by modifying a component along with its abstracted concepts (see Table~\ref{tab:mars_component_statistics_breakdown}), we first study whether exposing LLMs to more conceptualizations and abstract knowledge can enhance their metaphysical reasoning capabilities. 
For this purpose, we select CANDLE~\cite{DBLP:journals/corr/abs-2401-07286} as the knowledge source, which is an automatically constructed knowledge base containing 382K conceptualizations of events and abstract inferential knowledge.
We first convert event-conceptualization pairs into the task format of metaphysical event discrimination and reformat commonsense inferential knowledge to align with the objectives of the metaphysical inference and transition discrimination tasks.
More details are in Appendix~\ref{appendix:CANDLE_implementation}.

Three backbone models are then fine-tuned separately on CANDLE and~\name{}. 
Another group is sequentially fine-tuned on CANDLE and then on~\name{}. 
All models are then evaluated on the testing set of~\name{}, with the results reported in Table~\ref{tab:transfer_Concept_to_MARs}.
From the results, a significant improvement is observed across all tasks when the models are sequentially fine-tuned on CANDLE and~\name{}, compared to solely fine-tuning on CANDLE or~\name{}.

These findings indicate that the transfer of conceptualizations and abstract knowledge from CANDLE effectively enhances the performance of LMs in metaphysical reasoning tasks. 
Since CANDLE is constructed by distilling from an LLM without human labor, this opens up a scalable and cost-efficient approach to improving the metaphysical reasoning capabilities of LLMs.

\subsubsection{Impact of Component Types}
\begin{figure*}[t]
     \centering
     \includegraphics[width=1\linewidth]{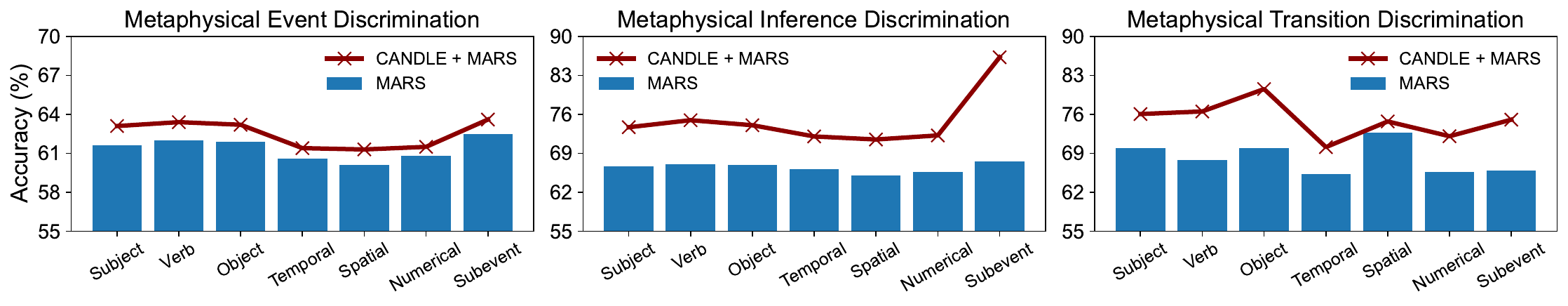}
     \vspace{-0.2in}
     \caption{Performances by component types of fine-tuned LLaMa3-8B on three tasks of~\name{}.}
    \label{fig:error_type_by_component}
    \vspace{-0.1in}
\end{figure*}

We then analyze the performance of LLMs on each component type to understand the reasons for their subpar performance. 
We select LLaMa-3-8B as the representative model and compare its accuracy on each component type when fine-tuned on~\name{} and CANDLE + \name{}.
The results are illustrated in Figure~\ref{fig:error_type_by_component}. 
We observe that while pre-training the model on CANDLE consistently enhances performance, LLaMa3 still struggles when reasoning with changes in spatial quantifiers, temporal quantifiers, and numerical properties. 
This is in line with recent studies that demonstrate weaknesses in temporal and numerical reasoning for LLMs~\cite{DBLP:conf/acl/TanNB23,DBLP:conf/icml/ShiCMSDCSZ23}. 
Another possible reason is that since CANDLE only contains conceptualizations for subjects, verbs, objects, and sub-events in social events, pre-training models on it cannot provide benefits for the aforementioned aspects of change. 
Moreover, we only observe limited improvement for the metaphysical event discrimination task.
Future works could focus on how to further enhance LLM's metaphysical reasoning capabilities in these weaker dimensions.

\subsubsection{Error Analysis of GPT-Series Models}
\label{sec:error_analysis}
Finally, we select GPT4 as a representative model and conduct a manual analysis to identify the causes of errors by categorizing the mistakes found in their COT responses. 
We sample 150 COT responses from each task, all of which result in inconsistent results compared to human annotated labels and present our classifications of these errors as follows:
\textbf{(1) Hallucinations}: 41.7\% of errors are caused by factual or metaphysical hallucinations by GPT4, where it creates a context that accommodates changes in actions and inferences that are not mentioned in the original text. 
For instance, in the event ``The poet enjoys writing poems about western festivals,'' GPT4 incorrectly interprets the poet as Du Fu. 
This leads to a conflict when reasoning about his life and the subsequent inference ``He was famous in the west,'' resulting in faulty  reasoning.
\textbf{(2) Confusion between Concepts and Hypernyms}: 36.3\% errors are attributed to GPT4's tendency to perceive abstract components within changed actions as hypernyms that fulfill the change, without considering all potential entities within the original concept. 
For instance, in a modified event, ``He jumps down from \textit{very high altitude} and lands peacefully,'' GPT4 interprets \textit{very high altitude} as a diving platform, deeming it plausible. 
However, this concept could also encompass high buildings, which would not be suitable for the event.
\textbf{(3) Internal Conflict}: 17.7\% errors are attributed to internal conflicts within GPT4's reasoning rationales, as well as inconsistencies between the binary predictions made and the corresponding reasoning rationales.
\textbf{(4) Annotation Error}: 4.3\% errors are erroneously identified due to incorrect labels, potentially caused by spamming or a misunderstanding of the task by human annotators.

\section{Conclusions}
\label{sec:conclusion}
In conclusion, this paper proposes \textbf{\textit{Metaphysical Reasoning}} to delineate the process of \textit{reasoning with changes in distribution} and construct \emojiname{} as the associated evaluation benchmark in a non-trivial manner.
Our experiments show the challenge of our task, which advanced prompting and fine-tuning can't easily solve. 
Analysis reveals why LMs struggle with metaphysical reasoning and suggests a possible improvement.
We hope to illuminate the path toward achieving conscious processing in LLMs through System II reasoning by effectively comprehending changes in distribution.

\newpage
\section*{Limitations}
\label{appendix:limitations}
Though we consider our work to be a fundamental step towards understanding the capabilities of LMs in \textit{reasoning with changes in distribution}, we do acknowledge that several limitations still exist that just cannot be covered within one single work. 
Here, we discuss some important limitations that future works can address:

\textbf{(1) Include more types of changes in our current formulation.}
In our work, we primarily focus on seven types of changes, covering the subject, verb, object, spatial quantifier, temporal quantifier, numerical properties, and sub-events of the event. 
While these seven types encompass most of the potential changes, there are other uncovered components within an event that can be impacted by changes, such as adjectives, adverbs, and prepositional phrases.
Nevertheless, our flexible and automated benchmark curation pipeline, empowered by an LLM, allows for future research to extend the benchmark to cover a broader range of component types.

\textbf{(2) Reliance of LLM on benchmark curation.}
Our data construction process relies significantly on ChatGPT, an expensive and proprietary language model used for data collection, as well as human annotation for data verification. 
In Appendix~\ref{appendix:replace_chatgpt_with_llama}, we discussed the feasibility of leveraging open-sourced LLM as a replacement to ChatGPT to reduce cost and promote reproducibility.
Future research could also consider utilizing robust open-source language models~\cite{DBLP:journals/corr/abs-2403-05530} and general statement plausibility estimators~\cite{DBLP:conf/emnlp/0010WWS0H23} to replace these methods. 

\textbf{(3) Solution and Downstream Applications of Metaphysical Reasoning.} 
While this paper establishes a comprehensive evaluation benchmark for metaphysical reasoning, we leave the exploration of a practical solution to aid LLMs in solving metaphysical reasoning tasks, as well as the potential benefits of utilizing metaphysical reasoning for downstream tasks into future works. 
These tasks may include planning~\cite{DBLP:conf/acl/YuanCFGSJXY23,DBLP:conf/emnlp/Ouyang023} or reasoning with changes~\cite{DBLP:conf/acl/0001HX023}.


\section*{Ethics Statement}
\label{appendix:ethics_statement}
\noindent\textbf{Offensive Content Elimination.}
Our benchmark curation pipeline, which involves generating content with ChatGPT, necessitates stringent measures to ensure the absence of offensive content in both the prompts and the generated responses. 
For this purpose, we apply two strategies to eliminate offensive content. 
First, we use the highest level of Azure AI Content Safety Filter to filter out any content that contains personal privacy, promotes violence, racial discrimination, hate speech, sexual content, or self-harm. 
If any such unsafe content is detected in the prompts or generated responses, it automatically triggers a system failure, which prevents the inclusion of such data in our dataset. 
Second, we manually inspect a random sample of 500 data entries from three tasks in~\emojiname{} for offensive content. 
Based on our annotations, we have not detected any offensive content.
We thus believe that our dataset is safe and will not yield any negative societal impact.

\noindent\textbf{Licenses.}
We will share our code and models under the MIT license, thereby granting other researchers free access to our assets for research purposes. 
Other datasets used in this paper, including Wikitext and Bookcorpus, are shared under the CC-SA license, permitting us to use them for research. 
As for language models, we access all open-source LMs via the Huggingface Hub~\cite{DBLP:conf/emnlp/WolfDSCDMCRLFDS20}. 
All associated licenses permit user access for research purposes, and we have agreed and committed to follow all terms of use.

\noindent\textbf{Annotations.}
We conduct large scale human annotations on the Amazon Mechanical Turk (AMT) platform. 
We invite annotation workers from the US, Europe, and India due to their proficiency in English.
The annotators are paid on average at an hourly rate of 19 USD, which is comparable to the minimum wages in the US. 
The selection of these annotators is solely based on their performance on the evaluation set, and we do not collect any personal information about the participants from AMT. 
For expert verifications, we have secured IRB approval and support from our institution's department, which allows us to invite expert graduate students to validate the quality of our data. 
They all agree to participate voluntarily without being compensated.
We have made concerted efforts to eliminate offensive content, thereby ensuring that no annotators are offended.

\section*{Acknowledgements}
We thank the anonymous reviewers and the area chair for their constructive comments. 
The authors of this paper were supported by the ITSP Platform Research Project (ITS/189/23FP) from the ITC of Hong Kong, SAR, China, as well as the AoE (AoE/E-601/24-N), the RIF (R6021-20), and the GRF (16205322) from the RGC of Hong Kong, SAR, China.

\bibliography{custom}

\newpage
\appendix

\begin{figure*}[t]
     \centering
     \includegraphics[width=1\linewidth]{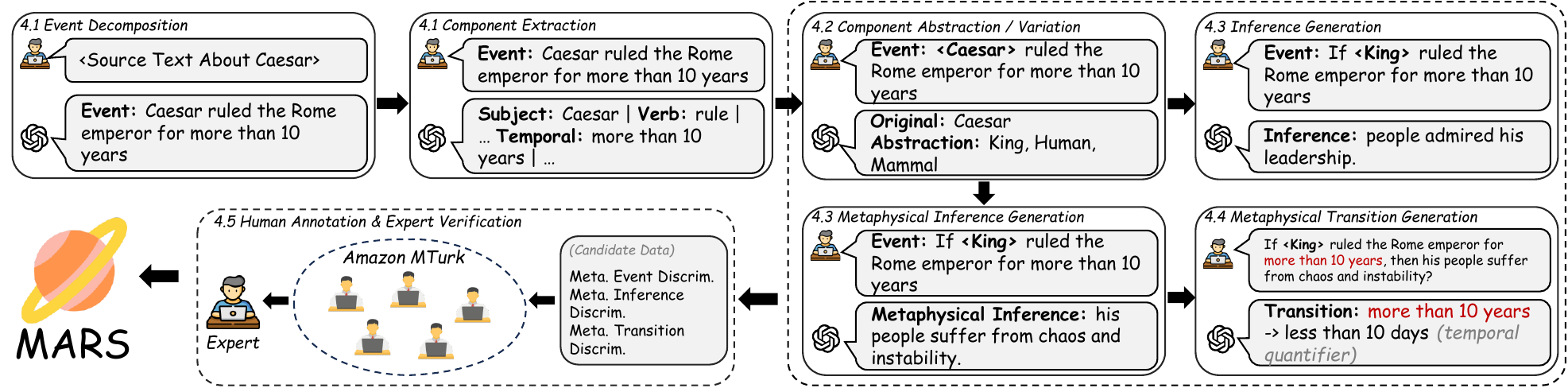}
     \caption{An overview of our benchmark curation pipeline with running examples.}
    \label{fig:data_curation_pipeline}
\end{figure*}

\begin{center}
    {\Large\textbf{Appendices}}
\end{center}

\section{Differentiation from Philosophical Metaphysics and Counterfactual Reasoning}
\label{appendix:metaphysical_versus_counterfactual}
In this work, we use the term ``metaphysical'' to describe a specific mode of reasoning that deals with highly improbable or abstract scenarios, distinct from both its traditional philosophical meaning and the concept of counterfactual reasoning.
Philosophically, ``metaphysics'' refers to the study of the fundamental nature of reality, encompassing questions about existence, causality, and the nature of being~\cite{aristotle1933metaphysics,bergson1999introduction}.
While this classical usage involves conceptual analysis and abstract thought, our focus diverges significantly.
We adopt ``metaphysical'' to signify reasoning that examines transitions between plausible and highly improbable states, emphasizing the logical structure and abstracted nature of these transitions rather than ontological or existential inquiries.

This distinction is important because our framework does not engage with the philosophical debates about the nature of reality or existence.
Instead, it concentrates on how LLMs process and adapt to scenarios that are rare or abstract yet logically consistent.
For example, while metaphysical reasoning in our context might involve reasoning about a scenario where ``a civilization survives for 100,000 years,'' it does not explore the metaphysical nature of time, existence, or causality in a philosophical sense.

Furthermore, our concept of metaphysical reasoning is distinct from counterfactual reasoning. 
Counterfactual reasoning involves evaluating ``what if'' scenarios that diverge from known realities but remain bounded by plausible causal relationships~\cite{DBLP:conf/acl/LiYE23,DBLP:conf/acl/HuaGDZNW24}.
For example, a counterfactual might consider, ``What if Caesar had lost the battle of Pharsalus?''--a scenario grounded in historical plausibility. 
In contrast, metaphysical reasoning in our framework extends beyond plausibility to explore scenarios that are structurally coherent but unlikely or abstract, such as ``What if Caesar ruled for a millennium?'' 
Here, the focus is not on causal plausibility but on the ability to evaluate transitions to rare, abstract, or highly improbable states.

This differentiation between ``metaphysical'' in our framework, metaphysics in philosophy, and counterfactual reasoning underscores the novel challenges our benchmarks aim to address.
By pushing LLMs to reason about transitions into abstract or improbable scenarios, we aim to probe and enhance their capabilities for adaptive, out-of-distribution reasoning -- a necessary step toward achieving generalizable System II reasoning.

\section{\name{} Benchmark Curation Details}
\label{appendix:prompts}
\subsection{\emojiname{} Benchmark Curation}
An overview of our benchmark construction pipeline is shown in Figure~\ref{fig:data_curation_pipeline}.
We first present our prompts used in each step for sequentially instructing ChatGPT to generate candidate data for~\emojiname{}~\cite{EcomScript}.

\subsubsection{Text Decomposition and Event Component Extraction}
To decompose a lengthy text from the source corpora into several action events, we use the following prompt to instruct ChatGPT.
\begin{displayquote}
\textbf{\texttt{\small 
You are required to decompose the given long sentence into several short yet semantically complete events, each describing an action.
An action event refers to those describing an action or a state change that occurs at a specific time and place.
The key components of each event should be preserved: including the subject, verb, object, temporal and spatial quantifiers, numerical properties of the subject and objects, and sub-events.
Generate one event as a whole sentence per line. You can generate as many events as you need. Below are some examples:
}}\\
\ldots \\
\textbf{\textcolor{headcolor}{\texttt{\small Sentence <i>: In November 2010, after years of planning and development, SpaceX successfully launched their Falcon 9 rocket into orbit for the first time. The launch took place at Cape Canaveral Air Force Station in Florida. The Falcon 9 carried a Dragon spacecraft mock-up, representing a major milestone in SpaceX's efforts to develop a reliable and cost-effective means of transporting cargo and eventually astronauts to the International Space Station.}}\\
\textcolor{tailcolor}{\texttt{\small\textbf{Event 1: SpaceX successfully launched their Falcon 9 rocket into orbit for the first time in November 2010.}}}\\
\textcolor{tailcolor}{\texttt{\small\textbf{Event 2: The Falcon 9 carried a Dragon spacecraft mock-up.}}}\\
\textcolor{tailcolor}{\texttt{\small\textbf{Event 3: The launch of the Falcon 9 took place at Cape Canaveral Air Force Station in Florida.}}}}\\
\ldots \\
\textbf{\textcolor{headcolor}{\texttt{\small Sentence <N>: In May 1934, following reports of a Japanese spy operating out of Dutch Harbor, the United States Navy dispatched Edwin T. Layton to the Aleutians to investigate the allegations.}}}
\end{displayquote}

We then use the following prompt to extract seven types of components from the decomposed events.
\begin{displayquote}
\textbf{\texttt{\small 
Given a short event, extract these components:\\
1. Subject: The noun that performs the action in the sentence.\\
2. Verb: The action word in the sentence.\\
3. Object: The noun that receives the action of the verb.\\
4. Temporal Quantifier: The time or time period of the event in the sentence.\\
5. Spatial Quantifier: The location or spatial extent of the event in the sentence.\\
6. Numerical Quantities and Properties of Objects: Numerical values describing the number or properties of the subject, object, or sub-events.\\
7. Sub-events: Complete events that are part of the main event in the sentence.\\
For each component, if there are more than one, separate them with |. If you cannot find one for a component, generate ``None'' only. Below are some examples:
}}\\
\ldots \\
\textbf{\textcolor{headcolor}{\texttt{\small Event <i>: After the First Battle of Naktong Bulge, the US Army's 2nd Infantry Division was moved to defend the Naktong River line.}}\\
\textcolor{tailcolor}{\texttt{\small\textbf{Subject: US Army's 2nd Infantry Division}}}\\
\textcolor{tailcolor}{\texttt{\small\textbf{Verb: moved | defend}}}\\
\textcolor{tailcolor}{\texttt{\small\textbf{Object: None}}}\\
\textcolor{tailcolor}{\texttt{\small\textbf{Temporal Quantifier: After the First Battle of Naktong Bulge}}}\\
\textcolor{tailcolor}{\texttt{\small\textbf{Spatial Quantifier: Naktong River line}}}\\
\textcolor{tailcolor}{\texttt{\small\textbf{Quantities and Properties of Objects: None}}}\\
\textcolor{tailcolor}{\texttt{\small\textbf{Sub-events: The US Army's 2nd Infantry Division was moved | The US Army's 2nd Infantry Division was moved to defend the Naktong River line.}}}}\\
\ldots \\
\textbf{\textcolor{headcolor}{\texttt{\small Event <N>: The University of Colorado created the Department of Medicine in September 1883 in the Old Main building on the Boulder campus.}}}
\end{displayquote}

\subsubsection{Component Abstraction and Variation}
For each type of component, we customize the prompt according to the nature of the component and whether the changes are implemented via abstraction or numerical variation. 
Here, we take the subject category with its abstraction as an example.
\begin{displayquote}
\textbf{\texttt{\small 
Given an event and a subject within the event, abstract the given subject in the given sentence into three different concepts. 
Each concept should be more abstract than the previous one. 
You are encouraged to be creative, but please ensure the three concepts gradually cover more instances. 
Below are some examples:
}}\\
\ldots \\
\textbf{\textcolor{headcolor}{\texttt{\small Event <i>: World's leading scientists announce breakthrough in clean energy technology, revolutionizing global sustainability efforts.}}\\
\textcolor{headcolor}{\texttt{\small Subject: World's leading scientists}}\\
\textcolor{tailcolor}{\texttt{\small\textbf{Concepts: expert, human, organism}}}}\\
\ldots \\
\textbf{\textcolor{headcolor}{\texttt{\small Event <N>: A driver is speeding down the highway.}}\\
\textcolor{headcolor}{\texttt{\small Subject: A driver}}}
\end{displayquote}

Note that leveraging LLM to perform contextualized abstraction~\cite{DBLP:journals/corr/abs-2401-07286,DBLP:conf/acl/YuWLBSLG0Y23} has been shown to result in better quality, larger coverage, and stronger downstream benefits compared to previous conceptualization methods~\cite{HE2024104149,DBLP:conf/acl/WangFXBSC23,DBLP:journals/corr/abs-2406-10885}, such as retrieving from a pre-defined concept taxonomy or human annotation.
Our knowledge distillation-based method is justifiable and enables large-scale benchmark construction.

\subsubsection{Inference Generation}
We use different prompts to collect plausible inferential states and metaphysical inferential states for each changed action event. 
Here, we provide the prompt for generating a metaphysical inference as an example.
\begin{displayquote}
\textbf{\texttt{\small 
Given an action event, generate a short metaphysical if-then inferential statement that describes an inferential state that only occurs in metaphysical space. 
A state is a condition or situation in which someone or something exists in the past or present that will last for a certain time if no changes occur.
An action is a thing that can be done in a time interval that is usually not long.
Metaphysical inference is a type of inference that is not based on empirical evidence but rather on the nature of things.
It can be a counterfactual inference that is contrary to the facts or reality, meaning that it is usually not true in reality world. 
Below are some examples:
}}\\
\ldots \\
\textbf{\textcolor{headcolor}{\texttt{\small Event <i>: In 2003, he played a recurring role on two episodes of The Bill.}}\\
\textcolor{tailcolor}{\texttt{\small Metaphysical Inference: Everyone criticizes his performance in the show.}}}\\
\ldots \\
\textbf{\textcolor{headcolor}{\texttt{\small Event <N>: Sam drives down the road with fast speed.}}}
\end{displayquote}

\begin{table*}[t]
\small
\centering
\resizebox{!}{!}{
\begin{tabular}{l|llllll}
\toprule
Model & Task 1 Plaus. & Expert. & Task 2 Plaus. & Expert. & Task 3 Plaus. & Expert. \\
\midrule
ChatGPT & 60.98 & 92.0 & 58.56 & 96.5 & 50.25 & 93.5 \\
Meta-LLaMa-3.1-405B & 62.2 & 93.2 & 57.0 & 95.8 & 51.0 & 94.6 \\
GPT-4o & 64.6 & 94.8 & 59.2 & 98.4 & 53.4 & 96.0 \\
\bottomrule
\end{tabular}
}
\caption{Annotation results of evaluation data curated with different LLMs as backbones. Plaus. refers to plausible event/inference/transition rate and Expert. refers to ratio of data accepted by expert annotators.}
\label{tab:LLM_comparison_for_data_construction}
\end{table*}

\subsubsection{Metaphysical Transition Generation}
Finally, we use the prompt below to collect the change needed to transition a metaphysical inference into a plausible one.
\begin{displayquote}
\textbf{\texttt{\small 
You will be given an event and its metaphysical inference, meaning that such an inference is impossible or rarely occurring in reality.
Please generate a transition that would make the inference plausible or possible in real life.
Specifically, you are required to only change a component of the event. The component must be one of the Subject, Verb, Object, Temporal Quantifier, Spatial Quantifier, Numerical Properties of Subject or Objects, and Sub-events of the event.
Below are some examples:
}}\\
\ldots \\
\textbf{\textcolor{headcolor}{\texttt{\small Event <i>: The boss of the company is monitoring the employees.}}\\
\textcolor{headcolor}{\texttt{\small Metaphysical Inference: The boss feels nervous and is expecting a rise.}}\\
\textcolor{tailcolor}{\texttt{\small\textbf{Transition: employees -> stocks (Object)}}}}\\
\ldots \\
\textbf{\textcolor{headcolor}{\texttt{\small Event <N>: The man is being chased by a 100 meters butterfly in the forest.}}\\
\textcolor{headcolor}{\texttt{\small Metaphysical Inference: The man is not scared and is laughing.}}}
\end{displayquote}

\begin{table*}[t]
\small
\centering
\resizebox{!}{!}{
\begin{tabular}{@{}l|ccc|c|ccc|c@{}}
\toprule
\multirow{2}{*}{Component Type} & \multicolumn{4}{c}{\textbf{Identified}} & \multicolumn{4}{c}{\textbf{Modified}} \\
\cmidrule(lr){2-5}\cmidrule(lr){6-9}
& ME. & MI. & MT. & \#Avg. & ME. & MI. & MT. & \#Avg. \\
\midrule
Subject & 4,376 & 3,907 & 3,507 & 1.116 & 3,106 & 2,950 & 2,591 & 1.094 \\
Verb & 9,874 & 8,856 & 8,061 & 3.647 & 4,408 & 4,146 & 3,760 & 3.457 \\
Object & 12,645 & 11,302 & 9,986 & 1.760 & 5,949 & 5,494 & 4,865 & 1.703 \\
Temporal Quantifier & 3,003 & 2,560 & 2,288 & 0.472 & 1,394 & 1,253 & 1,110 & 0.435 \\
Spatial Quantifier & 3,866 & 3,741 & 3,301 & 0.459 & 2,064 & 1,979 & 1,718 & 0.476 \\
Numerical Properties & 5,619 & 4,932 & 4,355 & 0.652 & 3,570 & 3,353 & 2,920 & 0.612 \\
Sub-events & 419 & 385 & 326 & 0.040 & 425 & 402 & 332 & 0.037 \\
\midrule
Total & 39,802 & 35,683 & 31,824 & 8.146 & 20,916 & 19,577 & 17,296 & 7.814 \\
\bottomrule
\end{tabular}
}
\caption{Number of unique components by type in annotated splits of~\name{}. \#Avg. refers to the average number of unique identified/modified component per event.}
\label{tab:mars_component_statistics_breakdown}
\end{table*}

\subsection{Main Evaluations on~\emojiname{}}
\label{appendix:prompt_main_evaluation}
To evaluate LLMs on three tasks in~\emojiname{}, we show our evaluating prompts in zero-shot scenario in Table~\ref{tab:appendix_baselines_prompt}.
Note that we are aware that LLMs may not be familiar with the word ``metaphysical.''
Therefore, we also experimented with replacing the word with ``implausible,'' and the best performances from both types of prompts are reported.
These models are consistent across all models' evaluations for fair comparison.

For few-shot evaluations, few shot examples are added after task descriptions and before the prompted test entry.
The exemplars are randomly sampled for each different test entry. 
For COT prompting, we specifically ask LLMs to ``think step by step and generate a short rationale to support your reasoning.''
Then, we ask it to give an answer based on its generated rationale.
The sampling temperature $\tau$ is set to 0.1 by default, and 5 COT responses are sampled with $\tau$ set to 0.7 in the SC-COT setting.

\subsection{Leveraging Open-sourced LLM for Benchmark Curation}
\label{appendix:replace_chatgpt_with_llama}
In this paper, we use proprietary LLMs and human annotation for data construction, which can be expensive and labor-intensive. 
However, this approach serves the best pursuit of data quality, which is crucial for an evaluation benchmark. 
Prior to our data collection, we tested a wide variety of LLMs, and ChatGPT outperformed almost all of them. 
Therefore, we opted to use it for data construction.
Nevertheless, with the recent advancements in state-of-the-art LLMs, we have found that \texttt{meta-llama/Llama-3.1-405B-Instruct} and \texttt{GPT-4o} also achieve satisfactory performance within our data collection framework. 
We sampled 500 original data entries and employed similar prompts and data collection processes to gather metaphysical reasoning evaluation data entries. 
We then asked expert annotators to rate the plausibility of the obtained data. 
The results are shown in Table~\ref{tab:LLM_comparison_for_data_construction}.
We observe that LLAMA3.1-405B can achieve comparable performance to ChatGPT in terms of plausible data (evaluation data that reflects reality rather than metaphysics, similar to the majority vote results in Table 2) and expert acceptance rates. 
Additionally, we find that GPT-4o can even improve the data collection process, resulting in higher quality data.
Thus, we believe this represents a compromise between data quality, reproducibility, and cost. 
It would also be feasible for data collectors to use LLAMA3.1 in the future for collecting metaphysical data, although leveraging proprietary LLMs can be more reliable to some extent.

\subsection{Additional Statistics on~\emojiname{}}
Table~\ref{tab:mars_component_statistics_breakdown} presents detailed statistics on the number of unique identified and modified components by type in the annotated splits of each task. 
The majority (approximately 80\%) of the components focus on the subject, verb, and object, while the remainder (around 20\%) concentrate on temporal quantifiers, spatial quantifiers, numerical properties, and sub-events.
On average, each annotated event in~\name{} features 8.15 identified components for changes and 7.81 transitions.

\begin{table*}[t]
\centering
\resizebox{\linewidth}{!}{
\begin{tabular}{@{}l|l@{}}
\toprule
Task & Prompt \\ 
\midrule
ME. & \begin{tabular}[c]{@{}l@{}}Given an event, determine whether it is a metaphysical event or not.\\A metaphysical event refers to event that is implausible or rarely occurring in reality.\\
If it is plausible and commonly accepted in the real world, answer yes.\\
On the contrary, if the event is metaphysical, answer No.\\
The event you need to discriminate is: \textbf{\texttt{<TEST-ENTRY-EVENT>}}.\\
Answer Yes or No only with one word:\end{tabular} \\ 
\midrule
MI. & \begin{tabular}[c]{@{}l@{}}Given an assertion that describes a if-then inference, determine whether the inference is plausible or metaphysical.\\
A plausible inference is an inference that is likely to be true or reasonable based on the information provided in the assertion.\\
A metaphysical inference is an inference that is not based on empirical evidence but rather on the nature of things,\\ it rarely occurs in the real world and can be counterfactual or implausible.\\
The assertion is: \textbf{\texttt{<TEST-ENTRY-INFERENCE>}}.\\
Answer Yes or No only with one word.\end{tabular}\\ 
\midrule
MT. & \begin{tabular}[c]{@{}l@{}}You are given an event, an inference based on the event that rarely occurs in the real world (a metaphysical inference),\\
and a transition in the event that would make the inference plausible or possible in the real world,\\
please determine whether the transition is correct or not in terms of making the inference plausible or possible.\\
The event is: \textbf{\texttt{<TEST-ENTRY-EVENT>}}.\\
The inference is: \textbf{\texttt{<TEST-ENTRY-INFERENCE>}}.\\
The transition is: \textbf{\texttt{<TEST-ENTRY-TRANSITION>}}.\\
Answer Yes or No only with one word.\end{tabular}\\
\bottomrule
\end{tabular}
}
\caption{Prompts used for evaluating LLMs across three tasks in~\emojiname{} in zero-shot scenario. 
ME. MI., and MT. stand for three tasks, respectively.}
\label{tab:appendix_baselines_prompt}
\end{table*}

\section{Implementation Details}
\label{appendix:implementation_details}
This section provides further implementation details for the main evaluations and subsequent analyses.

For all experiments, we use the Huggingface\footnote{\href{https://huggingface.co/}{https://huggingface.co/}} Library~\cite{DBLP:conf/emnlp/WolfDSCDMCRLFDS20} to build all models. 
For each LLM, we conduct experiments with both its instruction fine-tuned version (if any) and the original version.
The one achieving higher performances will be included in the reported results.
For LLaMa2, the model code is \texttt{meta-llama/Llama-2-7b/13b/70b(-chat)-hf}.
For LLaMa3, the model code is \texttt{meta-llama/Meta-Llama-3-8B/70B(-Instruct)}.
For Mistral, we use \texttt{mistralai/} \texttt{Mistral-7B(-Instruct)-v0.3}.

For ChatGPT and GPT4, we access it through Microsoft Azure APIs\footnote{\href{https://azure.microsoft.com/en-us/products/ai-services/}{https://azure.microsoft.com/en-us/products/ai-services/}}.
The code of the accessed version for ChatGPT is \texttt{gpt-35-turbo}, and for GPT4 is \texttt{gpt-4}.
Both models are of the version dated \texttt{2024-02-01}.
The maximum generation length is set to 50 tokens in zero-shot and few-shot settings, while for COT and SC-COT evaluations, the maximum generation length is set at 200 tokens. 

All experiments are conducted on eight NVIDIA-V100 (32G) GPUs, with 8E disk space, 48 CPU cores, and 1T memory. 
Each experiment is repeated three times with different random seeds, and the average performances are reported.
The variance across all experiments remains below 0.08, which is considered extremely small. 
Due to space constraints, we omit reporting this variance.

\subsection{Main Evaluations on~\emojiname{}}
\label{appendix:implementation_main_evaluations}
First, we add random voting and majority voting as another two baselines for revealing the characteristics of the~\emojiname{} benchmark.

To evaluate PTLMs in a zero-shot manner, we adopt the evaluation pipeline used for zero-shot question answering~\cite{DBLP:conf/aaai/MaIFBNO21,DBLP:conf/emnlp/WangF0XLSB23}.
Specifically, we convert each discrimination data entry into two declarative statements, which serve as natural language assertions corresponding to `yes'' or ``no'' options. 
For instance, when determining whether an event is metaphysical, we generate two assertions: ``The event \texttt{<EVENT>} is metaphysical as it's unlikely to occur in reality,'' and ``The event \texttt{<EVENT>} is not metaphysical; it's plausible in reality.'' 
The models are then tasked with computing the loss of each assertion. 
The assertion with the lowest loss is considered as the model's prediction. 
This approach allows any PTLM to be evaluated under classification tasks with an arbitrary number of options or even type classification based on a single assertion. 
We use the open code library\footnote{\href{https://github.com/Mayer123/HyKAS-CSKG}{https://github.com/Mayer123/HyKAS-CSKG}} as our code base and follow the default hyperparameter settings.
For VERA, we follow the exact same implementation\footnote{\href{https://github.com/liujch1998/vera}{https://github.com/liujch1998/vera}}~\cite{DBLP:conf/emnlp/0010WWS0H23}.
The accessed backbone model is \texttt{liujch1998/vera}, and all other hyperparameter settings follow the default implementation.

For fine-tuning PTLMs, we connect each PTLM backbone with five fully connected classification layers. 
The entire model is then fine-tuned using a classification objective with cross-entropy loss. 
We employ a default setting of a learning rate of 5e-6 and a batch size of 64. 
The models are optimized using an AdamW optimizer~\cite{DBLP:conf/iclr/LoshchilovH19}, with the model's performance evaluated every 50 steps. 
We set the maximum sequence lengths for the tokenizers to 70 for all three discriminative subtasks. 
Early stopping is also implemented to select the best checkpoint when the highest validation accuracy is achieved. 
To ensure convergence, we train all models with five epochs.

For evaluating LLMs in a zero-shot manner, we transform the input for each task into assertions using natural language prompts, as illustrated in Table~\ref{tab:appendix_baselines_prompt}.
The models are then prompted to determine the plausibility of the provided assertions by answering yes or no questions.
We parse their responses using pre-defined rules to derive binary predictions.
When generating each token, we consider the top 10 tokens with the highest probabilities.

For fine-tuning LLMs, we use LoRA for fine-tuning, and the LoRA rank and $\alpha$ are set to 16 and 32, respectively.
We adopt the open code library from LlamaFactory\footnote{\href{https://github.com/hiyouga/LLaMA-Factory}{https://github.com/hiyouga/LLaMA-Factory}}~\cite{DBLP:journals/corr/abs-2403-13372} for model training and evaluation.
We similarly use an Adam~\cite{DBLP:journals/corr/KingmaB14} optimizer with a learning rate of 5e-5 and a batch size of 8.
The maximum sequence length for the tokenizer is set at 300.
All models are fine-tuned over three epochs, selecting the checkpoint with the highest accuracy on the validation set.

Finally, for evaluating proprietary LLMs, such as ChatGPT and GPT4, we similarly prompt them as with open LLMs.
Detailed prompts are explained in Appendix~\ref{appendix:prompt_main_evaluation}.

We also include full evaluation results (with more baselines and models included) in Table~\ref{tab:main_eval_full_results}.
Specifically, for RAG~\cite{DBLP:journals/corr/abs-2312-10997}, we reformulate the traditional paradigm of retrieval-augmented generation for our task by asking an LLM to first identify important concepts from the evaluation data entry, retrieve relevant knowledge from an abstract knowledge base containing information about the concepts, and merge them into the evaluation prompt for making the final prediction on metaphysical reasoning tasks. 
This approach aligns with the design of our~\name{} benchmark and provides insights into which method offers more benefits when comparing retrieval to fine-tuning conceptual knowledge into LLMs.

For Multi-Agent Calibration, we adopt the multi-agent deliberation design from~\citet{DBLP:journals/corr/abs-2404-09127}, which is a multi-agent confidence calibration system for multiple-choice QA. 
In this setting, we set up two LLMs. The first LLM generates the initial chain-of-thought response and prediction for each task. 
The second LLM is prompted with the first LLM's chain-of-thought response and is asked to analyze the differences. 
Its reasoning rationale regarding these differences, particularly in the metaphysical realm, is then provided as feedback to the first LLM. 
The first LLM incorporates this feedback and is asked to regenerate the chain-of-thought rationale and final prediction. This loop continues until the second LLM agrees with the first LLM.

For Self-Reflection~\cite{DBLP:journals/tacl/PanSXNWW24}, we adopt a straightforward approach to rectify LLM errors by using feedback provided by the LLM itself (self-reflection). 
In this setting, we first ask an LLM to generate a chain-of-thought response explaining the rationale behind a given metaphysical data entry. We then prompt it for a new round, deliberately asking it to analyze the correctness of its rationale and answer. 
This feedback is merged back into the original prompt and first response to generate a refined response after self-reflection.

\subsection{Improving Metaphysical Reasoning via Transferring from Conceptualization Taxonomy}
\label{appendix:CANDLE_implementation}
In this section, we elaborate further on how we transform CANDLE into the format of three tasks in~\emojiname{} for large-scale pre-training in improving LMs' metaphysical reasoning abilities.

CANDLE's data is primarily divided into two sections.
The first section comprises conceptualizations of instances or events, which can be reformatted into metaphysical event discrimination. 
Each data entry in CANDLE represents a conceptualization of an abstracted instance within an event or the abstraction of an entire event. 
Following our definition in Section~\ref{sec:task_definitions}, we interpret each conceptualization as a change in the event.
For each data entry, replacing the original instance with its conceptualization forms a plausible change that could occur in reality. 
Subsequently, we randomly select negative conceptualizations for an event from conceptualizations of other events that do not share any common words with the anchor event.
These negative conceptualizations form metaphysical events. 
Three models are then pre-trained on four million events, with a balanced ratio of plausible events and metaphysical events.
The hyperparameters for fine-tuning all models remain consistent with the implementation details described above in Appendix~\ref{appendix:implementation_main_evaluations}.

The second part contains the commonsense inferential knowledge of abstracted events, which can be interpreted as inferential states of the modified events. 
To synchronize with our task structure, we exclusively select relations that imply a state in the inferential knowledge. 
We obtain negative inference samples in a similar manner by sampling from inference tails of events without common keywords. 
Subsequently, we pre-train models for both the metaphysical inference discrimination task and the metaphysical transition reasoning task. 
These models are trained to determine whether the inference is plausible or metaphysical in relation to the altered event. 
As CANDLE does not include transitions, this approach serves as the most accurate simulation of the metaphysical transition reasoning task.
It's also important to note that CANDLE is exclusively predicated on social events, covering only subject, object, and sub-events as types of abstraction changes.
In contrast,~\emojiname{} contains a significantly wider array of events, incorporates more types of changes, and also evaluates (L)LMs' capabilities in discerning what additional change is requisite to instigate a transition.
These features make~\emojiname{} distinct from tasks in CANDLE.

\section{Annotation Details}
\label{appendix:annotation_details}
\subsection{Worker Selection Protocol}
To ensure the high quality of our human annotation, we implement strict quality control measures.
Initially, we invite only those workers to participate in our qualification rounds who meet the following criteria: 1) a minimum of 1K HITs approved, and 2) an approval rate of at least 95\%. 
We select workers separately for each task and conduct three qualification rounds per task to identify those with satisfactory performance. 
In each qualification round, we create a qualification test suite that includes both easy and challenging questions, each with a gold label from the authors. 
Workers are required to complete a minimum of 20 questions. 
To qualify, they must achieve an accuracy rate of at least 80\% on the qualification test. 
After our selection process, we chose 36, 24, and 32 workers for three tasks, respectively, from a pool of 481, 377, and 409 unique annotators. 
On average, our worker selection rate stands at 7.26\%. 
Following the qualification rounds, workers are required to complete another instruction round. 
This round contains complex questions selected by the authors, and workers are required to briefly explain the answer to each question. 
The authors will then double-check the explanations provided by the annotators and disqualify those with a poor understanding.

\subsection{Annotation Interface}
\label{appendix:annotation_interface}
\begin{figure*}[t]
     \centering
     \includegraphics[width=1\linewidth]{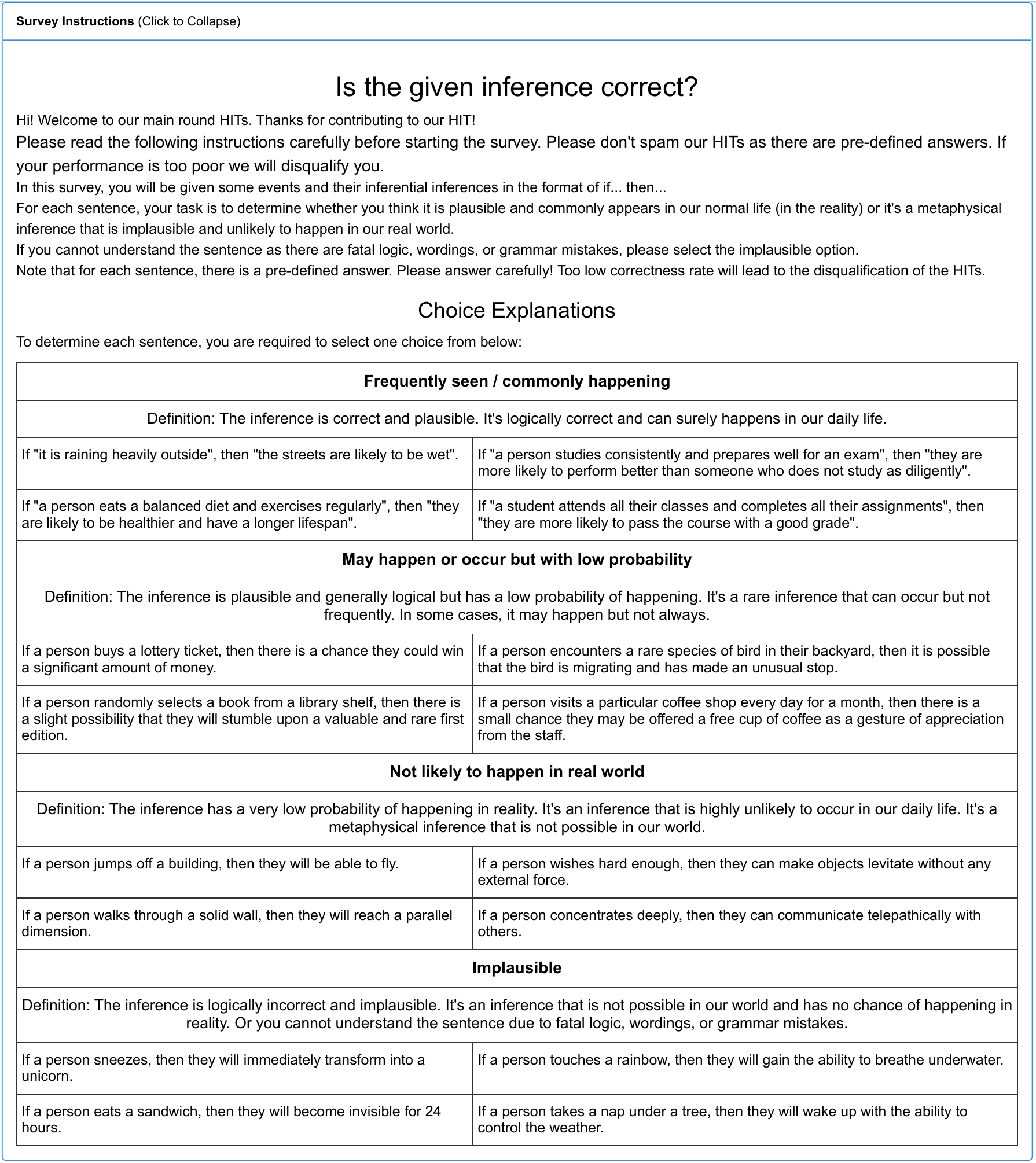}
     \caption{Our annotation instruction for the workers at the metaphysical inference discrimination task.
     Workers are provided with both task explanations and detailed examples.}
    \label{fig:annotation_instruction}
\end{figure*}

\begin{figure*}[t]
     \centering
     \includegraphics[width=0.7\linewidth]{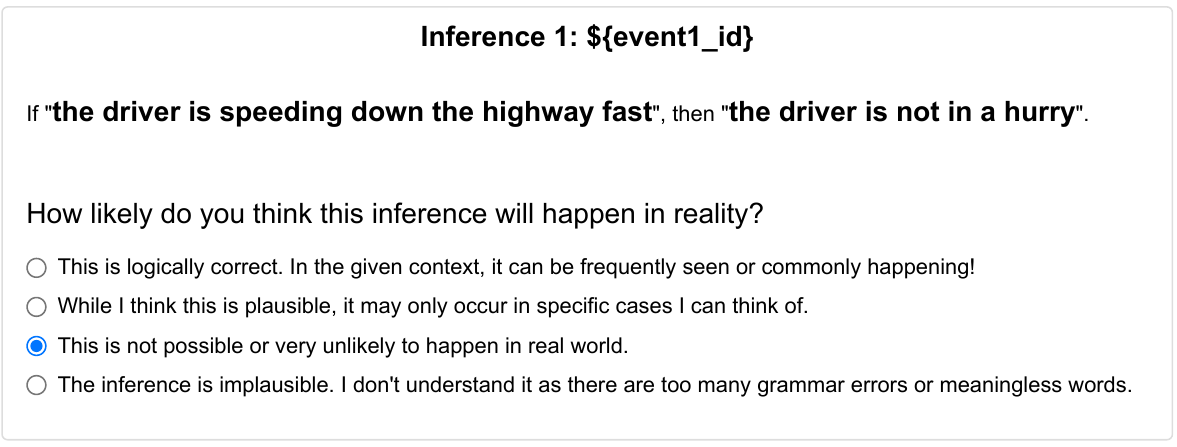}
     \caption{An example of a question that has been released to the worker. Workers are asked to annotate in a four-point Likert scale.}
    \label{fig:Question_demo}
\end{figure*}

For each task, we provide workers with comprehensive task explanations in layman's terms to enhance their understanding. 
We also offer detailed definitions and several examples of each choice to help annotators understand how to make decisions.
Each entry requires the worker to annotate using a four-point Likert scale. 
Workers are asked to rate the plausibility of the given question using such scale, where 1 signifies strong agreement and 4 indicates strong disagreement. 
We consider annotations with a value of 1 or 2 as plausible and those with a value of 3 or 4 as implausible. 
A snapshot of our annotation instructions, along with a snapshot showing the question released to the worker, are shown in Figure~\ref{fig:annotation_instruction} and Figure~\ref{fig:Question_demo}. 
To ensure comprehension, we require annotators to confirm that they have thoroughly read the instructions by ticking a checkbox before starting the annotation task. 
We also manually monitor the performance of the annotators throughout the annotation process and provide feedback based on common errors. 
Spammers or underperforming workers will be disqualified.
The overall inter-annotator agreement (IAA) stands at 81\% in terms of pairwise agreement, and the Fleiss kappa~\cite{fleiss1971measuring} is 0.56. 
These statistics are generally comparable to or slightly higher than those of other high-quality dataset construction works~\cite{DBLP:conf/aaai/SapBABLRRSC19,DBLP:conf/emnlp/FangWCHZSH21,DBLP:conf/www/FangZWSH21,DBLP:conf/aaai/HwangBBDSBC21,DBLP:journals/corr/abs-2502-16169,DBLP:conf/nips/BaiLW0S23}, which indicates that the annotators are close to achieving a strong internal agreement.

\subsection{Expert Verification}
Finally, we enlist the help of three postgraduate students, each with extensive experience in NLP research, to validate the annotations. 
These students are given the same instructions as those provided to the crowd-sourcing workers and are asked to verify a sample of 100 annotations for each task. 
The high level of consistency between our expert annotators and the AMT annotators, as demonstrated in Table~\ref{tab:mars_statistics}, suggests that our AMT annotation is of high quality.

\section{Additional Experiments and Analysis}
\label{appendix:additional_experiments}
In this section, we include additional analytical experiments to provide better support for our claims in~\name{}.

\subsection{Multi-task Fine-tuning on~\emojiname{}}
\label{appendix:multitask-finetuning}
\subsubsection{Setup}
To achieve conscious processing, an ideal language model should be capable of performing three tasks uniformly and sequentially. 
However, fine-tuning each task separately contradicts this objective, as it results in a model that can only perform one task after one training. 
Therefore, in this section, we investigate the possibility of enabling a language model to master all tasks simultaneously through multitask fine-tuning.
Given that all three tasks are binary classification tasks, we adopt a straightforward approach. 
The language model is trained using a randomly shuffled combination of training data from all three tasks. 
This anticipates that the model will learn all tasks collectively. 
The best checkpoint is chosen based on achieving the highest accuracy on the validation sets of all three tasks. 
After training, the model performance is evaluated separately on the testing sets of each task. 
All training details remain consistent with those explained in the Appendix~\ref{appendix:implementation_main_evaluations}.

\subsubsection{Results and Analysis}
The results are presented in Table~\ref{tab:multitask_eval_results}. 
Upon analyzing these results, we observe that LLMs fine-tuned in a multi-task setting generally outperform those simply fine-tuned on the respective training data for each task. 
This observation is interesting as it suggests that training the model uniformly across all three tasks can enhance the entire process simultaneously, thereby improving reasoning with changes in distribution. 
This implies that LLMs can potentially mimic human learning abilities, which are better equipped to reason with changes by collectively understanding the feasibility, consequence, and necessity of such changes. 
Such a phenomenon indirectly indicates that our task formulation is indeed interconnected and collectively forms a reasoning pipeline. 
However, it's important to note that this improvement is only marginal. 
LLMs still exhibit limited metaphysical reasoning ability, particularly in the metaphysical event discrimination task.
More advanced methods are still required to enable LLMs to achieve metaphysical reasoning.

\subsection{Few-shot Fine-tuning on~\emojiname{}}
\label{appendix:few-shot}
\subsubsection{Setup}
From the main evaluation results in Table~\ref{tab:main_eval_test}, it is evident that fine-tuning consistently enhances the performance of all models on~\emojiname{}.
In this section, we delve deeper into the impact of fine-tuning in a few-shot setting, with the aim of analyzing the performance of models trained with limited data. 
More specifically, we aim to examine how models perform with varying sizes of training data. 
This will enable us to determine whether collecting more data invariably benefits fine-tuning, thereby leading to the development of more robust metaphysical reasoners. 
To achieve this, we sample the training data for each task in a progressively increasing ratio of 0.2, 0.4, 0.6, 0.8, and 1.0, and use each sampled training data to fine-tune LLMs for each task individually. 
The models are then evaluated on the complete validation sets to select the optimal checkpoint, and on the full testing set for performance assessment. 
All fine-tuning parameters remain consistent across all models, as detailed in Appendix~\ref{appendix:implementation_main_evaluations}.

\begin{table*}[t]
    \small
	\centering
    \resizebox{!}{!}{
\begin{tabular}{l|l|ccc}
\toprule
\textbf{Data Split}   & \textbf{Evaluation Method} & \textbf{Event-ACC} & \textbf{Inference-ACC} & \textbf{Transition-ACC} \\ 
\midrule
MARS                  & Zero-shot                 & 53.90              & 51.20                  & 49.41                   \\ 
MARS                  & Few-shot                  & 49.85              & 51.47                  & 48.88                   \\
MARS-Claude           & Zero-shot                 & 54.50              & 54.00                  & 53.50                   \\
MARS-Claude           & Few-shot                  & 56.00              & 55.50                  & 54.00                   \\
MARS-LLAMA3.1         & Zero-shot                 & 52.00              & 56.50                  & 56.50                   \\
MARS-LLAMA3.1         & Few-shot                  & 55.50              & 57.50                  & 58.50                   \\
\bottomrule
\end{tabular}
    }
    \caption{Evaluation results (\%) of GPT-4o on~\name{} constructed with different backbone LLMs.} 
    \label{tab:chatgpt_bias_eval}
\end{table*}

\subsubsection{Results and Analysis}
The results are reported in Table~\ref{tab:fewshot_MARS}.
From these results, we observe that training the model with a few-shot training data sample generally has a negative impact across all tasks in~\emojiname{}. 
However, this impact is not significant, and on rare occasions, the sampled training data even leads to superior results compared to training on the full sets. 
When the training data is reduced to different ratios (80\%, 60\%, 40\%, and 20\%), the performance of the models is not significantly affected. 
This suggests that the models are capable of learning from a small amount of training data and that performance is not significantly influenced by the size of the training data. 
In other words, annotating more data for training does not necessarily result in better performance, indicating that our task cannot be simply resolved by increasing training data. 
Future research can explore more advanced reasoning paradigms or training methods to further enhance the capabilities of LLMs in metaphysical reasoning.

\subsection{Fine-tuned PTLMs vs. Fine-tuned LLMs}
\label{appendix:ptlms_vs_llms}
To validate the reason why fine-tuned PTLMs perform better than fine-tuned LLMs, we first hypothesis that PTLMs have a faster convergence rate to the training data due to their smaller number of parameters and fully fine-tuned paradigm (compared to LoRA when fine-tuning LLMs).
This results in better fine-tuned performance than LLMs.
Although LLMs have lower performance, they exhibit stronger generalizability to other tasks.
We fine-tune a DeBERTa-v3 model with 25\% and 50\% of the training data and observed their performance in Table~\ref{tab:fewshot_MARS}.
From the results, we observe that when we reduce the training data for PTLMs, they are hardly comparable to fine-tuned LLMs. However, the last 50\% of randomly sampled data brought significant improvements.
While we cannot determine the exact reason due to the black box nature of these language models, we believe that PTLMs have a faster rate of fitting into the distribution of the training data or human annotations, resulting in better outcomes on human-annotated evaluation sets.
LLMs are more likely to learn how to make correct inferences rather than simply fitting the data.
Another possible reason is that we use LoRA to fine-tune LLMs due to limited computational resources; fully fine-tuning LLMs might further enhance their performance.

\subsection{Inherent Bias in~\name{} Construction}
\label{appendix:ChatGPT_bias}
One concern regarding the~\name{} benchmark is the potential bias introduced by using GPT-series models, specifically ChatGPT, for dataset construction. 
Our approach to constructing~\name{} was guided by the need to balance scalability with quality. 
In pilot studies evaluating metaphysical reasoning across various models, GPT-series models consistently demonstrated the highest levels of creativity and reliability. 
Based on these findings, we selected GPT as the primary backbone for data generation. 
Constructing~\name{}, however, required extensive manual annotation, as LLMs often fail to provide accurate labels for complex reasoning tasks. 
This manual verification process made it impractical to create multiple versions of~\name{} using different backbone LLMs due to expensive human labors required.
Thus, to address concerns about potential biases arising from reliance on ChatGPT, we conducted additional experiments by constructing two smaller versions of the~\name{} benchmark. 
These alternative benchmarks utilized data generated from two different LLMs, Claude-3.5-sonnet~\cite{claude} and LLAMA 3.1-70B~\cite{LLAMA3}, in each step, to obtain 200 evaluation data entries per task in~\name{}.
All samples underwent expert annotation to collect ground-truth labels. 
We then evaluate GPT-4’s zero-shot and few-shot performance on these alternative benchmarks alongside the original~\name{}.

The results are shown in Table~\ref{tab:chatgpt_bias_eval}.
We observe that using different LLMs as backbones for~\name{} construction results in similar performance by GPT-4 across zero-shot and few-shot settings. 
Overall, the difficulty of the~\name{} benchmark remains robust and consistent, irrespective of the backbone LLM used during dataset generation. 
These experiments demonstrate that the reliance on ChatGPT for the original~\name{} construction does not compromise the benchmark’s validity or difficulty. 
The results reinforce the reliability of MARS as a comprehensive test of metaphysical reasoning, with its complexity surpassing any potential biases introduced by the specific LLM used in data collection.

\subsection{Binary Task Design in~\name{}}
\label{appendix:TaskDesign}
In~\name{}, all tasks are designed as a binary prediction task to facilitate automated and easy label collection and evaluation.
Here, we discuss the reason and some pilot analysis behind such task design by considering other task formulations, including multiple-choice, open-ended generation, and binary evaluation. 

Multiple-choice tasks, while structured and amenable to automated evaluation, posed significant challenges in collecting high-quality negative (distractor) options. 
Relying on human annotators to create distractors proved labor-intensive and impractical for scaling, as it required drafting multiple plausible but incorrect options for each question.
As a result, we adopted open-ended generation and binary evaluation, ultimately choosing a generate-then-annotate paradigm. 
This approach involved two stages: first, evaluating the performance of LLMs in generating metaphysical cases during the generation phase; second, annotating the generated cases with binary labels (correct/incorrect). 

To complement the binary evaluation results, we also included human annotation results for ChatGPT’s performance in generating metaphysical data, as indicated in the \textit{Majority} row of Table~\ref{tab:main_eval_test}, which can be regarded as following a generative task paradigm.
The results demonstrate that, even when the task is framed as a generation task, ChatGPT struggles with metaphysical reasoning. 
The low proportion of human-annotated correct generations highlights the difficulty of reasoning about metaphysical changes, regardless of task formulation. 
While binary evaluation offers clear performance metrics and scalability advantages, the generation task provides complementary insights into the model’s creative and reasoning capabilities. 
Together, these observations underscore the importance of improving LLMs' ability to reason about distributional and situational changes, which is crucial for advancing their metaphysical reasoning capabilities.

\section{Case Studies}
\label{appendix:case_studies}
In this section, we present some examples for each of the three tasks in~\emojiname{} to help readers better understand our benchmark. 
The examples are displayed in Table~\ref{tab:case_study_mars}. 
We observe that examples in~\emojiname{} typically require careful reasoning and consideration of the plausibility of occurrences in reality or the metaphysical realm to make the correct discrimination.

\clearpage

\begin{table*}[t]
    \small
    \centering
    \resizebox{0.88\linewidth}{!}{
	\begin{tabular}{@{}llccccccccc@{}}
	\toprule
    \multirow{2}{*}{\textbf{Methods}}&\multirow{2}{*}{\textbf{Backbone}}&\multicolumn{3}{c}{\textbf{Event}} &\multicolumn{3}{c}{\textbf{Inference}}&\multicolumn{3}{c}{\textbf{Transition}}\\
    \cmidrule(lr){3-5}\cmidrule(lr){6-8}\cmidrule(lr){9-11}
	&&\textbf{Acc}&\textbf{AUC}&\textbf{Ma-F1}&\textbf{Acc}&\textbf{AUC}&\textbf{Ma-F1}&\textbf{Acc}&\textbf{AUC}&\textbf{Ma-F1}\\
            \midrule
            \textbf{Random} & \multicolumn{1}{c}{-} & 50.00 & - & 49.56 & 50.00 & - & 49.56 & 50.00 & - & 49.56 \\
            \textbf{Majority} & \multicolumn{1}{c}{-} & 60.98 & - & 37.99 & 58.56 & - & 36.93 & 50.25 & - & 33.37 \\
		  \midrule
            \multirow{8}{*}{\textbf{\begin{tabular}[c]{@{}l@{}}PTLM\\ \textit{(Zero-shot)}\end{tabular}}}
            &RoBERTa-Base \scriptsize{\textit{211M}} & 38.60 & 49.40 & 27.90 & 44.30 & 55.11 & 30.80 & 51.13 & 53.37 & 38.36 \\
            &RoBERTa-Large \scriptsize{\textit{340M}} & 38.57 & 50.94 & 27.83 & 44.37 & 56.49 & 30.73 & 50.90 & 53.08 & 33.92 \\
            &DeBERTa-Base \scriptsize{\textit{214M}} & \underline{60.55} & 49.41 & 42.89 & 50.10 & 47.57 & 48.96 & 49.05 & 41.32 & 33.19 \\
            &DeBERTa-Large \scriptsize{\textit{435M}} & 48.27 & 49.88 & 45.87 & 47.73 & 49.94 & 44.44 & 50.73 & 46.96 & 46.15 \\
            &GPT2-XL \scriptsize{\textit{1.5B}} & 38.62 & \underline{51.12} & 27.93 & 44.40 & 51.88 & 31.45 & 49.92 & 48.35 & 48.09 \\
            &CAR \scriptsize{\textit{435M}} & 54.63 & 49.34 & 49.96 & 48.33 & 42.85 & 41.93 & 52.97 & 35.05 & 46.94 \\
            &CANDLE \scriptsize{\textit{435M}} & 51.90 & 49.12 & \underline{50.30} & 46.77 & 44.03 & 38.48 & 53.49 & 34.95 & 47.95 \\
            &VERA \scriptsize{\textit{11B}} & 51.82 & 50.48 & 48.52 & \underline{60.97} & \underline{62.54} & \underline{59.09} & \underline{61.31} & \underline{66.32} & \underline{61.17} \\
            \midrule
            \multirow{6}{*}{\textbf{\begin{tabular}[c]{@{}l@{}}PTLM\\ \textit{(Fine-tuned)}\end{tabular}}}
            &RoBERTa-Base \scriptsize{\textit{211M}} & 63.32 & 62.76 & 61.76 & 69.08 & 70.54 & 68.90 & 71.24 & 72.73 & 70.65 \\ 
            &RoBERTa-Large \scriptsize{\textit{340M}} & 64.22 & 63.18 & 62.62 & 69.04 & 70.63 & 68.90 & 69.68 & 71.70 & 68.73 \\ 
            &DeBERTa-Base \scriptsize{\textit{214M}} & 63.82 & 63.98 & \textbf{\underline{63.39}} & 69.50 & 70.59 & 69.31 & 71.96 & 73.85 & 71.17 \\ 
            &DeBERTa-Large \scriptsize{\textit{435M}} & \textbf{\underline{64.45}} & \textbf{\underline{64.16}} & 63.27 & \textbf{\underline{69.57}} & \textbf{\underline{71.15}} & 69.33 & \textbf{\underline{72.93}} & 74.00 & 72.01 \\
            &GPT2-XL \scriptsize{\textit{1.5B}} & 46.68 & 47.63 & 46.96 & 43.70 & 44.22 & 30.41 & 44.57 & 45.03 & 45.89 \\ 
            &VERA \scriptsize{\textit{11B}} & 61.95 & 61.43 & 60.81 & 63.90 & 66.93 & \textbf{\underline{70.84}} & 71.75 & \textbf{\underline{74.57}} & \textbf{\underline{73.27}} \\
            \midrule
            \multirow{15}{*}{\textbf{\begin{tabular}[c]{@{}l@{}}LLM\\ \textit{(Zero-shot)}\end{tabular}}}
            & Meta-LLaMa-2-7B & 50.64 & - & 41.41 & 49.87 & - & 49.23 & 50.94 & - & 50.64 \\
            & Meta-LLaMa-2-13B & 51.50 & - & 49.48 & 50.81 & - & 50.57 & 50.81 & - & 50.80 \\
            & Meta-LLaMa-2-70B & 52.40 & - & 49.03 & 56.13 & - & 46.81 & 48.45 & - & 48.34 \\
            & Meta-LLaMa-3-8B & 50.62 & - & 49.12 & 51.33 & - & 50.98 & 51.95 & - & 51.07 \\
            & Meta-LLaMa-3-70B & 57.41 & - & 50.59 & 63.40 & - & 61.82 & 60.15 & - & 60.01 \\
            & Meta-LLaMa-3.1-8B & 51.01 & - & 50.27 & 52.13 & - & 51.29 & 52.35 & - & 52.09 \\
            & Meta-LLaMa-3.1-70B & 59.22 & - & 52.08 & 63.61 & - & 61.90 & 61.28 & - & 61.03 \\
            &+RAG & \underline{61.21} & - & \underline{54.51} & \underline{66.38} & - & \underline{65.90} & 61.53 & - & 61.22 \\
            &+Multi-Agent & 56.12 & - & 51.08 & 65.06 & - & 65.01 & \underline{62.54} & - & \underline{62.19} \\
            &+Self-reflection & 57.94 & - & 53.17 & 63.91 & - & 63.51 & 60.92 & - & 60.77 \\
            & Meta-LLaMa-3.1-405B & 59.22 & - & 52.08 & 63.61 & - & 61.90 & 61.28 & - & 61.03 \\
            & Gemma-2-9B & 56.88 & - & 48.53 & 51.83 & - & 51.76 & 49.41 & - & 45.01 \\
            & Falcon-7B & 54.32 & - & 49.51 & 51.77 & - & 50.30 & 50.42 & - & 49.02 \\
            & Falcon-40B & 52.35 &  - & 50.36 & 49.67 & - & 49.38 & 50.27 & - & 50.22 \\
            & Mistral-7B & 49.90 & - & 48.94 & 50.23 & - & 50.06 & 51.75 & - & 51.75 \\
            \midrule
            \multirow{5}{*}{\textbf{\begin{tabular}[c]{@{}l@{}}LLM\\ \textit{(Fine-tuned)}\end{tabular}}}
            & Meta-LLaMa-2-7B & 60.10 & 59.90 & 59.00 & 63.51 & 66.44 & 62.55 & 66.06 & 70.38 & 65.12 \\
            & Meta-LLaMa-2-13B & 60.67 & 60.64 & 60.00 & 64.61 & 67.67 & 63.59 & 68.22 & 72.19 & 66.37 \\
            & Meta-LLaMa-3-8B & 60.06 & 60.54 & 59.58 & 65.76 & 67.88 & 65.72 & 69.83 & 74.59 & 68.74 \\
            & Gemma-2-9B & \underline{61.23} & \underline{61.25} & \underline{60.28} & \underline{69.24} & \underline{70.76} & \underline{69.00} & \underline{73.30} & \underline{76.91} & \underline{69.18} \\
            & Mistral-7B & 60.35 & 60.77 & 60.07 & 66.91 & 70.06 & 65.95 & 71.87 & 75.47 & 68.53 \\
            \midrule
            \multirow{12}{*}{\textbf{\begin{tabular}[c]{@{}l@{}}LLM\\ \textit{(API)}\end{tabular}}}
            &ChatGPT & 51.00 & - & 50.35 & \underline{61.35} & - & \underline{57.63} & 60.40 & - & \underline{60.12} \\
            &ChatGPT (5-shots) & 53.61 & - & 53.28 & 58.05 & - & 57.42 & \underline{62.40} & - & 59.35 \\
            &ChatGPT (COT) & 53.20 & - & 52.61 & 50.40 & - & 50.32 & 49.95 & - & 49.83\\
            &ChatGPT (SC-COT) & 53.98 & - & 53.47 & 52.47 & - & 51.99 & 51.25 & - & 51.13\\
            &GPT4 & 53.90 & - & 53.45 & 51.20 & - & 50.95 & 49.41 & - & 49.33 \\
            &GPT4 (5-shots) & 49.85 & - & 49.58 & 51.47 & - & 51.30 & 48.88 & - & 48.73 \\
            &GPT4 (COT) & 51.28 & - & 50.73 & 51.49 & - & 51.35 & 47.62 & - & 47.58 \\
            &GPT4 (SC-COT) & 51.97 & - & 51.26 & 52.05 & - & 52.27 & 48.24 & - & 48.11 \\
            &GPT-4o-mini & 57.94 & - & 57.91 & 53.84 & - & 53.53 & 48.06 & - & 48.06 \\
            &+RAG & \underline{59.99} & - & \underline{59.97} & 54.54 & - & 54.21 & 49.39 & - & 49.19 \\
            &+Multi-Agent & 54.21 & - & 53.17 & 52.76 & - & 52.26 & 46.94 & - & 46.70 \\
            &+Self-reflection & 56.89 & - & 55.21 & 53.22 & - & 53.20 & 48.51 & - & 48.45 \\
		\bottomrule
	\end{tabular}
    }
\caption{Full evaluation results (\%) of various language models on the testing sets of~\name{}. The best performances within each method are \underline{underlined} and the best among all methods are \textbf{bold-faced}.} 
\label{tab:main_eval_full_results}
\end{table*}

\clearpage

\begin{table*}[t]
    \small
	\centering
    \resizebox{!}{!}{
	\begin{tabular}{@{}llccccccccc@{}}
	\toprule
    \multirow{2}{*}{\textbf{Methods}}&\multirow{2}{*}{\textbf{Backbone}}&\multicolumn{3}{c}{\textbf{Event}} &\multicolumn{3}{c}{\textbf{Inference}}&\multicolumn{3}{c}{\textbf{Transition}}\\
    \cmidrule(lr){3-5}\cmidrule(lr){6-8}\cmidrule(lr){9-11}
	&&\textbf{Acc}&\textbf{AUC}&\textbf{Ma-F1}&\textbf{Acc}&\textbf{AUC}&\textbf{Ma-F1}&\textbf{Acc}&\textbf{AUC}&\textbf{Ma-F1}\\
            \midrule
            \textbf{Random} & \multicolumn{1}{c}{-} & 50.00 & - & 49.56 & 50.00 & - & 49.56 & 50.00 & - & 49.56 \\
            \textbf{Majority} & \multicolumn{1}{c}{-} & 60.98 & - & 37.99 & 58.56 & - & 36.93 & 50.25 & - & 33.37 \\
            \midrule
            \multirow{9}{*}{\textbf{\begin{tabular}[c]{@{}l@{}}LLM\\ \textit{(Zero-shot)}\end{tabular}}}
            & Meta-LLaMa-2-7B & 50.64 & - & 41.41 & 49.87 & - & 49.23 & 50.94 & - & 50.64 \\
            & Meta-LLaMa-2-13B & 51.50 & - & 49.48 & 50.81 & - & 50.57 & 50.81 & - & 50.80 \\
            & Meta-LLaMa-2-70B & 52.40 & - & 49.03 & 56.13 & - & 46.81 & 48.45 & - & 48.34 \\
            & Meta-LLaMa-3-8B & 50.62 & - & 49.12 & 51.33 & - & 50.98 & 51.95 & - & 51.07 \\
            & Meta-LLaMa-3-70B & 57.41 & - & 50.59 & 63.40 & - & 61.82 & 60.15 & - & 60.01 \\
            & Gemma-1.1-7B & 56.88 & - & 48.53 & 51.83 & - & 51.76 & 49.41 & - & 45.01 \\
            & Falcon-7B & 54.32 & - & 49.51 & 51.77 & - & 50.30 & 50.42 & - & 49.02 \\
            & Falcon-40B & 52.35 &  - & 50.36 & 49.67 & - & 49.38 & 50.27 & - & 50.22 \\
            & Mistral-7B & 49.90 & - & 48.94 & 50.23 & - & 50.06 & 51.75 & - & 51.75 \\
            \midrule
            \multirow{5}{*}{\textbf{\begin{tabular}[c]{@{}l@{}}LLM\\ \textit{(Fine-tuned)}\end{tabular}}}
            & Meta-LLaMa-2-7B & 60.10 & 59.90 & 59.00 & 63.51 & 66.44 & 62.55 & 66.06 & 70.38 & 65.12 \\
            & Meta-LLaMa-2-13B & 60.67 & 60.64 & 60.00 & 64.61 & 67.67 & 63.59 & 68.22 & 72.19 & 66.37 \\
            & Meta-LLaMa-3-8B & 60.06 & 60.54 & 59.58 & 65.76 & 67.88 & 65.72 & 69.83 & 74.59 & 68.74 \\
            & Gemma-1.1-7B & 61.23 & 61.25 & 60.28 & 69.24 & 70.76 & 69.00 & 73.30 & 76.91 & 69.18 \\
            & Mistral-7B & 60.35 & 60.77 & 60.07 & 66.91 & 70.06 & 65.95 & 71.87 & 75.47 & 68.53 \\
            \midrule
            \multirow{5}{*}{\textbf{\begin{tabular}[c]{@{}l@{}}LLM\\ \textit{(Multi-task)}\end{tabular}}}
            & Meta-LLaMa-2-7B & 60.70 & 59.88 & 59.17 & 66.15 & 64.67 & 64.34 & 70.40 & 70.89 & 70.20 \\
            & Meta-LLaMa-2-13B & 61.36 & 61.42 & 60.69 & 67.07 & 66.44 & 65.68 & 70.44 & 69.15 & 68.62 \\
            & Meta-LLaMa-3-8B & 61.38 & 61.85 & 61.02 & 67.20 & 67.13 & 66.60 & 71.64 & 72.06 & 71.12 \\
            & Gemma-1.1-7B & 61.54 & 62.36 & 61.15 & 67.71 & 67.60 & 66.98 & 73.12 & 72.82 & 71.89 \\
            & Mistral-7B & 61.03 & 61.16 & 60.38 & 67.69 & 67.20 & 66.16 & 72.34 & 72.52 & 71.78 \\
		\bottomrule
	\end{tabular}
    }
    \caption{Evaluation results (\%) of LLMs fine-tuned on~\emojiname{} under the multi-task setting.} 
    \label{tab:multitask_eval_results}
\end{table*}

\begin{table*}[t]
\small
\centering
\resizebox{!}{!}{
\begin{tabular}{@{}l|c|ccccccccc@{}}
\toprule
    \multirow{2}{*}{\textbf{Backbone}}&\multirow{2}{*}{\textbf{Training Data}}&\multicolumn{3}{c}{\textbf{Event}} &\multicolumn{3}{c}{\textbf{Inference}}&\multicolumn{3}{c}{\textbf{Transition}}\\
    \cmidrule(lr){3-5}\cmidrule(lr){6-8}\cmidrule(lr){9-11}
    &&\textbf{Acc}&\textbf{AUC}&\textbf{Ma-F1}&\textbf{Acc}&\textbf{AUC}&\textbf{Ma-F1}&\textbf{Acc}&\textbf{AUC}&\textbf{Ma-F1}\\
\midrule
\multirow{5}{*}{\textbf{\begin{tabular}[c]{@{}l@{}}LLaMa-2\\ \scriptsize{\textit{7B}}\end{tabular}}} 
& 20\% & 58.03 & 58.24 & 57.62 & 62.43 & 64.47 & 60.43 & 63.11 & 63.08 & 62.73 \\
& 40\% & 58.81 & 58.40 & 57.69 & 64.03 & 67.48 & 61.58 & 66.44 & 70.04 & 64.15 \\
& 60\% & 59.09 & 59.41 & 58.62 & 64.75 & 68.10 & 62.79 & 67.00 & 70.85 & 64.15 \\
& 80\% & 59.48 & 60.54 & 59.82 & 64.15 & 68.01 & 61.53 & 66.42 & 70.64 & 64.92 \\
& 100\% & 60.10 & 59.90 & 59.00 & 63.51 & 66.44 & 62.55 & 66.06 & 70.38 & 65.12 \\
\midrule
\multirow{5}{*}{\textbf{\begin{tabular}[c]{@{}l@{}}LLaMa-2\\ \scriptsize{\textit{13B}}\end{tabular}}} 
& 20\% & 59.95 & 59.75 & 58.57 & 63.80 & 66.86 & 61.80 & 64.11 & 68.73 & 64.08 \\
& 40\% & 59.45 & 59.18 & 58.25 & 65.49 & 68.98 & 63.54 & 68.52 & 71.61 & 64.82 \\
& 60\% & 60.19 & 59.46 & 58.92 & 65.90 & 69.59 & 64.18 & 68.24 & 72.17 & 65.59 \\
& 80\% & 60.24 & 60.05 & 59.43 & 65.99 & 69.70 & 64.27 & 68.35 & 72.43 & 65.97 \\
& 100\% & 60.67 & 60.64 & 60.00 & 64.61 & 67.67 & 63.59 & 68.22 & 72.19 & 66.37 \\
\midrule
\multirow{5}{*}{\textbf{\begin{tabular}[c]{@{}l@{}}LLaMa-3\\ \scriptsize{\textit{8B}}\end{tabular}}} 
& 20\% & 60.56 & 59.91 & 58.99 & 63.40 & 66.77 & 61.06 & 65.23 & 70.50 & 64.60 \\
& 40\% & 60.68 & 59.98 & 59.23 & 62.35 & 69.00 & 61.81 & 69.43 & 72.72 & 65.27 \\
& 60\% & 60.74 & 60.88 & 60.49 & 65.90 & 69.59 & 61.81 & 69.00 & 72.78 & 65.55 \\
& 80\% & 60.91 & 61.03 & 60.29 & 66.73 & 69.71 & 61.72 & 68.71 & 73.15 & 66.43 \\
& 100\% & 60.06 & 60.54 & 59.58 & 65.76 & 67.88 & 65.72 & 69.83 & 74.59 & 68.74 \\
\midrule
\multirow{5}{*}{\textbf{\begin{tabular}[c]{@{}l@{}}Gemma-v1.1\\ \scriptsize{\textit{7B}}\end{tabular}}} 
& 20\% & 59.07 & 59.54 & 59.18 & 64.70 & 70.42 & 62.43 & 68.41 & 73.64 & 67.08 \\
& 40\% & 60.79 & 59.93 & 59.72 & 62.80 & 70.57 & 62.26 & 69.83 & 73.91 & 62.18 \\
& 60\% & 59.26 & 60.31 & 59.25 & 67.83 & 70.22 & 60.56 & 70.68 & 74.56 & 66.98 \\
& 80\% & 59.31 & 59.32 & 58.73 & 64.03 & 70.77 & 63.73 & 69.66 & 73.51 & 67.05 \\
& 100\% & 61.23 & 61.25 & 60.28 & 69.24 & 70.76 & 69.00 & 73.30 & 76.91 & 69.18 \\
\midrule
\multirow{5}{*}{\textbf{\begin{tabular}[c]{@{}l@{}}Mistral-v1.1\\ \scriptsize{\textit{7B}}\end{tabular}}} 
& 20\% & 60.67 & 60.27 & 59.61 & 65.28 & 69.22 & 63.16 & 68.37 & 72.85 & 66.15 \\
& 40\% & 60.53 & 60.78 & 60.03 & 65.92 & 70.21 & 63.96 & 69.79 & 72.97 & 69.46 \\
& 60\% & 61.82 & 61.86 & 61.07 & 67.65 & 70.46 & 64.09 & 67.92 & 73.38 & 66.76 \\
& 80\% & 59.35 & 59.55 & 58.85 & 68.07 & 70.43 & 66.49 & 69.84 & 73.63 & 65.84 \\
& 100\% & 60.35 & 60.77 & 60.07 & 66.91 & 70.06 & 65.95 & 71.87 & 75.47 & 68.53 \\
\midrule
\multirow{3}{*}{\textbf{\begin{tabular}[c]{@{}l@{}}DeBERTa-v3-Large\\ \scriptsize{\textit{635M}}\end{tabular}}} 
& 25\% & 58.11 & 57.90 & 57.64 & 63.28 & 64.12 & 64.70 & 64.51 & 67.21 & 66.54 \\
& 50\% & 61.32 & 59.71 & 60.91 & 65.36 & 67.12 & 68.09 & 67.95 & 68.21 & 67.97 \\
& 100\% & 64.45 & 64.16 & 63.27 & 69.57 & 71.15 & 69.33 & 72.93 & 74.00 & 72.01 \\
\bottomrule
\end{tabular}
}
\caption{Evaluation results (\%) of LLMs fine-tuned on~\emojiname{} under the few-shot setting. Training data refers to the ratio of sampled training data from the full training sets of~\emojiname{}.}
\label{tab:fewshot_MARS}
\end{table*}

\clearpage

\clearpage

\begin{table*}[t]
\small
\centering
{\def\arraystretch{1.4}
\begin{tabularx}{\textwidth}{>{\raggedright}p{0.5cm}>{\RaggedRight\arraybackslash}p{11.5cm}p{0.5cm}} \toprule
Task & Data Examples & Label \\ 
\midrule
ME. & The tax offices were devastation \original{burnt down} & \texttt{P.} \\
ME. & Keith and Vinnie are running \original{competition} against each other in the sheriff's election & \texttt{P.} \\
ME. & We worked together environment \original{in the marina} for years & \texttt{M.} \\
ME. & The sun is melting horizon \original{over the landscape} like an orange popsicle & \texttt{M.} \\
ME. & Mammal \original{human} seek food for their own survival & \texttt{P.} \\
\midrule
MI. & If I perception \original{felt} the tension leave me, then I feel more relaxed now & \texttt{P.} \\
MI. & If they both reached the excellence \original{world top 100} in 2005, then they both worked hard to achieve their goals & \texttt{P.} \\
MI. & If Parker and Garbajosa were adaptable \original{two very versatile players} who could both defend and attack, then they were actually terrible basketball players. & \texttt{M.} \\
MI. & If Stevens success \original{won} his first eight games, then Steven is a skilled player. & \texttt{P.} \\
MI. & If I communication \original{have to talk} to my insurance company, then my insurance company is not responsive and does not provide good customer service. & \texttt{M.} \\
\midrule
MT. & If he was respectful \original{overpowering and right intrusion}, then he will apologize for his actions and make amends. & \texttt{P.} \\
MT. & If the other guests have just been invited to participate in a karaoke session \original{join community on the dance floor}, then the other guests decline the invitation and choose to sit and watch instead. & \texttt{P.} \\
MT. & If Australia opposed \original{supported} South Vietnam in that time period, then Australia support South Vietnam during that time period. & \texttt{M.} \\
MT. & If Churchill has ignoring \original{communication} to the requests for verification in various ways, then Churchill is not interested in verifying the requests and is avoiding them. & \texttt{P.} \\
MT. & If Tikal has hundreds \original{thousands} of history structures, then archaeologists have not yet discovered the true purpose of Tikal's structures. & \texttt{M.} \\
\bottomrule
\end{tabularx}
}
\caption{Case studies of three tasks in the~\emojiname{} benchmark.
ME, MI, and MT refer to three tasks in metaphysical reasoning, respectively.
\texttt{P.} refers to plausible in reality and \texttt{M.} refers to metaphysical.
The original component before the change/transition is marked in \original{grey}.
}
\label{tab:case_study_mars}
\end{table*}

\end{document}